\documentclass[10pt,journal]{IEEEtran}

\usepackage{comment}
\usepackage{graphicx}
\usepackage{amsmath}
\usepackage{amssymb}
\usepackage{dsfont}
\usepackage{array}

\usepackage{color}
\usepackage{soul}

\newcolumntype{P}[1]{>{\centering\arraybackslash}p{#1}}

\usepackage{subcaption}
\usepackage{booktabs}
\usepackage{multirow}

\usepackage{setspace} 

\usepackage{url}
\usepackage[colorlinks=true, linkcolor=black, urlcolor=blue]{hyperref}
 
\begin{document}
\title{Uncertainty Quantification of Spatiotemporal Travel Demand with Probabilistic Graph Neural Networks}

\author{Qingyi Wang, Shenhao Wang$^*$, Dingyi Zhuang, Haris Koutsopoulos, Jinhua Zhao

\thanks{Qingyi Wang, Dingyi Zhuang, Jinhua Zhao: Massachusetts Institute of Technology, Cambridge, MA, 02139 USA.}
\thanks{Shenhao Wang: University of Florida, Gainesville, FL 32611 USA; Massachusetts Institute of Technology, Cambridge, MA, 02139 USA}
\thanks{Haris Koutsopoulos: Northeastern University, Boston, MA, 02115 USA.}
\thanks{Corresponding to: Shenhao Wang; e-mail: shenhaowang@ufl.edu}}


\maketitle

\begin{abstract}

Recent studies have significantly improved the prediction accuracy of travel demand using graph neural networks. However, these studies largely ignored uncertainty that inevitably exists in travel demand prediction. To fill this gap, this study proposes a framework of probabilistic graph neural networks (Prob-GNN) to quantify the spatiotemporal uncertainty of travel demand. This Prob-GNN framework is substantiated by deterministic and probabilistic assumptions, and empirically applied to the task of predicting the transit and ridesharing demand in Chicago. 
We found that the probabilistic assumptions (e.g. distribution tail, support) have a greater impact on uncertainty prediction than the deterministic ones (e.g. deep modules, depth). Among the family of Prob-GNNs, the GNNs with truncated Gaussian and Laplace distributions achieve the highest performance in transit and ridesharing data. 
Even under significant domain shifts, Prob-GNNs can predict the ridership uncertainty in a stable manner, when the models are trained on pre-COVID data and tested across multiple periods during and after the COVID-19 pandemic. 
Prob-GNNs also reveal the spatiotemporal pattern of uncertainty, which is concentrated on the afternoon peak hours and the areas with large travel volumes. 
Overall, our findings highlight the importance of incorporating randomness into deep learning for spatiotemporal ridership prediction.
Future research should continue to investigate versatile probabilistic assumptions to capture behavioral randomness, and further develop methods to quantify uncertainty to build resilient cities.

\end{abstract}


\section{Introduction}
\noindent
Uncertainty prevails in urban mobility systems. An enormous number of uncertainty sources range from the long-term natural disasters and climate change, to real-time system breakdown, and to the inherent randomness in travel demand. These uncertainty sources exhibit spatial and temporal patterns: travel demand could drastically change during the holiday seasons, or becomes spatially concentrated in the special events (e.g. football games). 
Traditionally, the spatial uncertainty is analyzed by spatial econometrics or discrete choice models, while the temporal one by time series models. 
Recently, deep learning models have been designed to capture the spatiotemporal correlations, with the Long Short Term Memory (LSTM) networks for temporal correlations and convolutional or graph neural networks (CNN, GNN) for spatial dependencies. Deep learning has been shown to significantly improve the prediction accuracy of travel demand. 
However, most studies focus on designing \textit{deterministic} deep learning models to predict the \textit{average} travel demand, but largely ignore the uncertainty inevitably associated with any prediction. 

Overlooking uncertainty leads to theoretical and practical problems. In theory, travel demand is inherently a random quantity, which is characterized by a rich family of probability distributions, sometimes even ``fat-tail'' ones that render a simple average approximation highly inappropriate \cite{Yang2019}. In practice, while the prediction of average demand provides a valuable basis for system design, the lack of uncertainty analysis precludes the opportunity of using deep learning to provide robust real-time service or resilient long-term planning \cite{Guo2021,guo2022data}. 
Since the recent work has demonstrated the outstanding capability of deep learning in modeling spatiotemporal dynamics, it seems timely and imperative to enrich the deterministic deep learning models to capture the spatiotemporal uncertainty of travel demand.

To address this research gap, we propose a framework of probabilistic graph neural networks (Prob-GNN) to quantify the spatiotemporal uncertainty in travel demand. 
This framework is substantiated by deterministic and probabilistic assumptions: the former refer to the various architectural designs in deterministic neural networks, while the latter refer to the probabilistic distributions with a variety of uncertainty characteristics.
The framework is used to compare the two types of assumptions by both uncertainty and point prediction metrics. 
This study makes the following contributions: 
\begin{enumerate}
    \item We identify the research gap in quantifying uncertainty for travel demand using deep learning. Despite a vast number of deep learning techniques for predicting average travel demand, only a limited number of studies have sought to quantify spatiotemporal uncertainty.
    \item To fill the research gap, the paper introduces a general framework of Prob-GNN, which comprises deterministic and probabilistic assumptions. This framework is further substantiated by twelve architectures, all of which can quantify the spatiotemporal uncertainty in travel demand.
    \item We find that probabilistic assumptions significantly influence uncertainty predictions, while only slightly influence point predictions. Conversely, the deterministic assumptions, such as graph convolution, network depth, and dropout rate, lead to a similar performance in both point and uncertainty predictions when probabilistic assumptions are fixed.
    \item We further demonstrate the generalizability of the Prob-GNNs by comparing the performance across multiple shifting domains caused by COVID-19. Although the mean prediction encounters significant prediction errors, the interval prediction is accurate across the domains. 
\end{enumerate}

\section{Literature Review}

The literature on spatiotemporal travel demand prediction using deep learning for has grown significantly in recent years. Among them, very limited number of studies have sought to understand the predictive uncertainty. This literature review will first examine the latest developments in demand predictions using deep learning, and subsequently, delve into uncertainty quantification. 

\subsection{Travel Demand Prediction with Deep Learning}
\noindent 
Researchers have achieved a lot of success in improving the prediction accuracy of spatiotemporal travel demand using advanced neural network architectural designs \cite{Yao2018,wu2020inductive,liu2022universal,zhuang2020compound}. 
Temporally, LSTM networks and Gated Recurrent Units (GRU) layers are used to analyze the seasonal, weekly, and time-of-day trends \cite{Ke_2017, Ye2020a, Liu2020}. 
Spatially, the analysis unit is often station/stop, urban grid, individuals, or census tract. The urban grid is often analyzed with convolutional neural networks (CNN) \cite{Yao2018}. Individual travel demand is analyzed by neural networks with behavioral insights for innovative architectural design \cite{wang2020_asu, wang2020_econ_info, wang2020_mtldnn}. More complex urban structures, such as transit lines that connect upstream and downstream stops, can be represented by graphs. Graph convolution networks (GCN) are then often used to model such spatial propagation of traffic/demand flow into adjacent nodes \cite{xiong2020dynamic, koca2021origin, Ye2020, liu2021deeptsp}. 
The baseline GCNs have been expanded by defining multiple graphs to describe the spatial similarity in points of interest (POIs) and transportation connectivity \cite{Geng2019,shi2020predicting}. An alternative to GCN is Graph Attention Network (GAT), which automatically learn the attentional weightings of each node, thus detecting the long-range dependencies without handcrafted weighting mechanisms through adjacency matrices \cite{liu2019attention,cao2021bert}. 

Despite the abundance of deep learning models, few studies sought to quantify the travel demand uncertainty, typically caused by narrowly defined prediction objectives and methods. The prediction objective - the average ridership - can hardly represent the randomness in travel demand, particularly for distributions with fat tails, in which the outliers deviating from the average value could predominate \cite{Yang2019}. 
The prediction method - the family of deterministic neural networks - is only a specific example of probabilistic neural networks with homogeneous assumptions on the variance. 
Demand uncertainty, if unaccounted for, can negatively influence downstream planning tasks \cite{guo2022data,wang2021uncertainty}. 
Uncertainty can also propagate and be significantly enhanced at each stage in multi-stage models \cite{Zhao2002}. Because of its importance, uncertainty has started to attract attention in the field of travel demand predictions using deep learning. 
Existing studies analyzed heteroskedastic noises in the Gaussian distribution \mbox{\cite{qian2023uncertainty,wangspatial}}, used Monte-Carlo dropouts to approximate model uncertainty\mbox{\cite{li2020,maryam2023uncertainty}}.
Additionally, the Gaussian assumption is being challenged. Studies modelled irregular distributions with quantile regressions \mbox{\cite{Rodrigues2020}}, and modelled sparse demand with zero-inflated distributions \mbox{\cite{zhuang2022uncertainty}}. 
Compared to the research in deterministic architectural designs, the current literature in predictive uncertainty in the travel demand realm using deep learning is still relatively thin. There lacks a comprehensive framework encompassing the deterministic and probabilistic assumptions to allow for a thorough comparison.

\subsection{Uncertainty Quantification Methods} \label{uncertainty_framework}
\noindent 
Research on uncertainty quantification has been developed in other deep learning fields. There are two major categories of uncertainty: data uncertainty and model uncertainty. Data uncertainty refers to the irreducible uncertainty inherent in the data generation process, while model uncertainty captures the uncertainties in the model parameters. For example, in linear regression, data uncertainty refers to the residuals, and model uncertainty refers to the standard errors of the estimated coefficients.

To characterize data uncertainty, either a parametric or non-parametric method can be used. 
Parametric methods refer to the models that parameterize a probabilistic distribution. Parametric models are often estimated by either Bayesian methods or mean-variance estimation (MVE).
Although conceptually appealing, Bayesian methods often involves unnecessarily intense computation, which relies on sampling methods and variational inference \cite{Pearce2018, Khosravi2011}. 
MVE minimizes the negative log-likelihood (NLL) loss based on a pre-specified distribution of the dependent variable \cite{Nix1994,khosravi2014optimized}. The MVE is computationally efficient, but it could lead to misleading results if probabilistic distributions are misspecified.
Non-parametric methods quantify uncertainty without explicitly imposing any parametric form. Typical examples include quantile regression \cite{Koenker2001} and Lower Upper Bound Estimation (LUBE) \cite{Pearce2018,Khosravi2011}. Non-parametric methods do not have misspecification problems, but it is hard to be optimized in complex neural networks. The pros and cons of both methods are systematically compared in review articles \cite{Kabir2018,Gawlikowski2021}.

Besides data uncertainty, uncertainty also emerges in model construction and training. Bayesian deep neural networks refer to the models that characterize all parameters of neural networks by probability distributions, typically Gaussian distribution \cite{zhuang2022uncertainty,Blundell2015}. 
A popular, low-cost way to do approximate Bayesian inference for neural networks with Gaussian parameters is to use dropouts \cite{li2020,Gal2016,Sicking2021}. 
Another source of model uncertainty comes from model training. Neural networks tend to converge to different local minima, resulting in different predictions. This type of uncertainty is usually addressed by bootstrapping or model ensembles \cite{Heskes1997, Lakshminarayanan2016}. Bootstrapping refers to the process of repeated training with re-sampling to capture the distributional uncertainty in sampling. Model ensembling refers to an average of multiple trainings procedures with different parameter initializations. 



Although the literature on travel demand uncertainty within the deep learning framework is scarce, classical statistics have decades of effort in quantifying travel demand uncertainty, with a focus on evaluating a rich family of probabilistic assumptions. Using high-resolution temporal ridership data, researchers adopted the ARIMA-GARCH model to quantify the volatility of subway demands during special events \cite{Chen2020} and adaptive Kalman filters to analyze real-time transit ridership \cite{Guo2014}. Using cross-sectional data, researchers used bootstrapping \cite{Petrik2016} to analyze parameter uncertainty, ensembling \cite{Cools2010} to analyze activity uncertainty, and heteroskedastic errors to describe the association between social demographics and ridership uncertainty in discrete choice models \cite{Petrik2016}. The common debates regarding the probabilistic assumptions include homoskedastic vs. heteroskedastic errors, Gaussian vs. exponential tails, real-line vs. positive support, and many others \cite{Chen2020,vlahogianni2011temporal,zhang2013univariate}, through which researchers could learn the detailed characteristics of travel demand uncertainty. The primary focus from the classical statistics on the probabilistic assumptions presents an interesting difference from the primary focus of the deep learning field in refining the deterministic assumptions. Indeed, it is quite ambiguous which set of assumptions is more crucial in quantifying the uncertainty of travel demand.

\section{Theoretical Framework}
\noindent
The authors propose a probabilistic graph convolutional neural network to compare the deterministic and probabilistic assumptions. The probabilistic GNN is represented as:
\begin{equation}
    y \sim \mathcal{G}(\theta) = \mathcal{G}(\mathcal{F}(\mathbf{X}, w))
\label{eq:meta_specification}
\end{equation}

\noindent
The first statement is a probability assumption about $y$, specified by the model family $\mathcal{G}(\theta)$, and the second uses the model family $\mathcal{F}(\mathbf{X}, w)$ to parameterize the probability distributions of $\mathcal{G}(\theta)$, with $\mathbf{X}$ being the inputs and $w$ being the model weights. While the recent studies focus on enriching $\mathcal{F}(\mathbf{X}, w)$ through spatiotemporal deep learning architectures, the potentially more critical probability assumption $\mathcal{G}(\theta)$ is largely neglected. The authors seek to compare the effects of the probabilistic assumptions $\mathcal{G}$, and the deterministic assumptions $\mathcal{F}$ in determining the model performance. 

Table \ref{tab:setup} summarizes the specifics of probabilistic and deterministic assumptions that substantiate the probabilistic GCN framework. A cross product of six probabilistic assumptions in $\mathcal{G}$ and two deterministic assumptions in $\mathcal{F}$ are tested, leading to twelve base models. Six probabilistic assumptions are: Homoskedastic Gaussian (HomoG), Poisson (Pois),  Heteroskedastic Gaussian (HetG), Truncated Gaussian (TG), Gaussian Ensemble (GEns), and Laplace (LAP). Two deterministic architectures, Graph Convolutional Networks (GCN) and Graph Attention Networks (GAT) are used to specify the probabilistic parameters. 

\begin{table*}[ht!]
\caption{Model Design. Upper panel: Probabilistic assumptions and their probability density functions. Bottom panel: Deterministic architecture and hyperparameters. A cross product of the two panels are used to substantiate the probabilistic GCNs, leading to twelve base models: HomoG-GCN, HomoG-GAT, HetG-GCN, HetG-GAT, TG-GCN, TG-GAT, GEns-GCN, GEns-GAT, Pois-GCN, Pois-GAT, Lap-GCN, and Lap-GAT.}
    \centering
    \resizebox{1.0\linewidth}{!}{%
    \begin{tabular}{l | c | c}
        \toprule
        \multicolumn{3}{l}{
        \textbf{Panel 1: Probabilistic Assumptions in $\mathcal{G}$}} \\
        \toprule
        Probabilistic assumptions & Probability density function & Distribution parameters $\theta$ \\
        \midrule
        Homoskedastic Gaussian (HomoG) & $f_{HomoG}(x;\mu,\sigma)=\frac{1}{\sigma\sqrt{2\pi}}\text{exp}(-\frac{1}{2}\frac{(x-\mu)^2}{\sigma^2})$, $\sigma = c$ & $\mu$ \\
        Poisson (Pois) & $f_{Pois}(x;\lambda)=\frac{\lambda^xe^{-\lambda}}{x!}$ & $\lambda$ \\
        Heteroskedastic Gaussian (HetG) & $f_{HetG}(x;\mu,\sigma)=\frac{1}{\sigma\sqrt{2\pi}}\text{exp}(-\frac{1}{2}\frac{(x-\mu)^2}{\sigma^2})$ & $\mu,\sigma$ \\
        Truncated Gaussian (TG) & $f_{TG}(x;\mu,\sigma) = \frac{f_G(x;\mu,\sigma)}{1-f_G(0;\mu,\sigma)}$ & $\mu,\sigma$ \\
        Gaussian Ensemble (GEns) & 
        $y_* \sim \mathcal{N}(\frac{1}{K} \sum_k{\mu_k}, \frac{1}{K} \sum_k{(\sigma_k^2 + \mu_k^2)} - \mu_*^2 )$ & $\mu_k,\sigma_k$ for $k=1..K$ models\\
        Laplace (Lap) & $f_{LAP}(x;\mu,b)=\frac{1}{2b}\text{exp}(-\frac{|x-\mu|}{b})$ & $\mu,b$ \\

        
        \midrule
        \midrule
        \multicolumn{3}{l}{\textbf{Panel 2: Deterministic Assumptions in $\mathcal{F}$}} \\
        \midrule
        Deterministic assumptions & Graph convolutional iteration function & Deterministic parameters $w$ \\
        \midrule
        Graph Convolutional Networks (GCN) \cite{Kipf2017} & $h^{(l+1)} = \sigma(\sum_{r=1}^R{\Tilde{D}_r^{-\frac{1}{2}}\Tilde{A}_r\Tilde{D}_r^{-\frac{1}{2}}h^{(l)}W_r^{(l)}})$ & $W_r^{(l)}$ \\
        Graph Attention Networks (GAT) \cite{velivckovic2017graph} & $h^{(l+1)} = \sigma(\sum_{j\in N(i)}{\alpha_{ij}Wh^{(l)}_j})$, $\alpha_{ij} = \frac{exp(W_ih^{(l)}_j)}{\sum_{k\in N_i}{exp(W_ih^{(l)}_k)}}$ & $\alpha_{ij}$, $W$ \\
        \midrule
        \multicolumn{3}{l}{Other hyperparameters in deterministic assumptions $\mathcal{F}$: Number of GCN/GAT/LSTM layers, Number of hidden layer neurons, Dropouts, Weight decay.} \\
        \bottomrule
    \end{tabular}
    }
    \label{tab:setup}
\end{table*}

\subsection{Probabilistic Assumptions $\mathcal{G}$}
\noindent 
The Gaussian distribution with a deterministic and homoskedastic variance term is chosen as the benchmark in specifying $\mathcal{G}$ because it is stable and has simple ensembling properties. This homoskedastic Gaussian assumption represents the vast number of deterministic deep learning models that use mean squared error as the training objective. The Gaussian benchmark facilitates the comparison across the probabilistic assumptions, as listed below.
\begin{enumerate}
    \item Homoskedasticity vs. heteroskedasticity \\
    We compare the heteroskedastic and homoskedastic Gaussian assumptions to examine how the data variance influences the model performance. The homoskedastic assumption assumes the same variance for all observations, whereas the heteroskedastic assumption estimates variance for every observation. It is highly likely that the travel demand variances has spatiotemporal patterns. 
    
    \item Continuous vs. discrete support\\
     We compare the Poisson distribution to the homoskedastic Gaussian benchmark to examine the effectiveness of continuous vs. discrete supports in determining the model performance. Since ridership takes integer values, the Poisson distribution could be more appropriate. However, the Poisson distribution uses only one parameter to represent both mean and variance, which could be overly restrictive.
    
    \item Real-line vs. non-negative support \\
    We compare the truncated heteroskedastic Gaussian distribution to the heteroskedastic Gaussian distribution to examine the effectiveness of the real-line vs. non-negative distribution support. Travel demand is non-negative, but the support of the Gaussian distribution covers the entire real line $(-\infty, +\infty)$. Therefore, a normal distribution left-truncated at zero is implemented to test whether non-negativity should be strictly imposed.
    
    \item Gaussian vs. exponential tails \\
    We compare the Laplace distribution to the heteroskedastic Gaussian distribution to examine whether the tail behavior of the distribution matters. The probability density function of the Gaussian distribution decays at a fast rate of $e^{x^2}$, while the Laplace distribution has a heavier tail with the decay rate of $e^{|x|}$.
    
    \item Single vs. ensembled models \\
    We compare the ensembled heteroskedastic Gaussian distributions to a single heteroskedastic Gaussian model to test whether an ensemble of distributions can outperform a single distributional assumption. The ensemble model is created by uniformly mixing $K$ estimates, which are trained independently with different parameter initializations. Suppose $\mathcal{Y}_k, \sigma_k^2$ are the estimated mean and variance for model $k$, and $\mathcal{Y}_*$ is the ensembled random variable. For the Gaussian distribution, the mixture can be further approximated as a Gaussian distribution, whose mean and variance are respectively the mean and variance of the mixture. The ensembled distribution is given by \cite{Lakshminarayanan2016}: $\mathcal{Y}_* \sim \mathcal{N}(\frac{1}{K} \sum_k{\mathcal{Y}_k}, \frac{1}{K} \sum_k{(\sigma_k^2 + \mathcal{Y}_k^2)} - \mathcal{Y}_*^2 )$.

\end{enumerate}

With the distributional assumption, maximum likelihood estimation (MLE) can be used to learn the parameters, which translates to minimizing the negative log-likelihood (NLL) loss in implementation. 
The NLL loss function based on the joint density of all observations is simply the negative sum of the log of the probability density of all observations:
\begin{equation}
    \text{NLL} = -\sum_{s,t} \text{log } \mathcal{G}(y_{st}|\mathbf{X_t}; w)
\end{equation}

In the uncertainty quantification literature, researchers typically differentiate between data (aleatoric) and model (epistemic) uncertainty \cite{Kiureghian2009}, and their sum is referred to as prediction uncertainty. The model uncertainty refers to the uncertainty resulting from the difference between $\mathcal{F}(\mathbf{X}, w)$ and the estimated $\hat{\mathcal{F}}(\mathbf{X}, w)$. The data uncertainty refers to the randomness in the data generation process and is represented by probabilistic assumptions $\mathcal{G}$, of which $y$ has the distribution. The prediction uncertainty combines model and data uncertainty, and describes the overall difference between the actual $y$ and the predicted $\hat{y}$. The relationship between the three quantities is directly given by $\sigma_y^2 = \sigma_{model}^2 + \sigma_{data}^2$. In our framework, assumptions 1-4 deal with the data uncertainty alone; while assumption 5 quantifies the prediction uncertainty.

\subsection{Deterministic Assumptions in $\mathcal{F}$}
\noindent
The deterministic assumptions in $\mathcal{F}$ consist of the spatial encoding, temporal encoding, and associated hyperparameter specifications. The following two subsections introduce the formulation of common spatial and temporal encodings. 

\subsubsection{GCN and GAT for Spatial Encoding}
\noindent
Two predominant spatial encoding methods - GCN and GAT - are adopted and compared to examine the effect of the deterministic architectures on model performance. The GCN layers need to access the global structure of the graph in the form of adjacency matrice(s), while the GAT layers aim to learn the spatial correlations from data. The propagation formula of both layers is introduced in Table \ref{tab:setup}.

To construct the multi-graph proximity in GCNs and GATs, four types of adjacency matrices are computed, including direct connectivity $[A_{Con}]_{ij} = 1 \text{ if two stations are adjacent, } =0 \text{ otherwise}$, network distance $[A_{Net}]_{ij} = \text{Network Distance}(i, j)^{-1}$, Euclidean distance $[A_{Euc}]_{ij} = \text{Euclidean Distance}(i, j)^{-1}$, and functional similarity $[A_{Func}]_{ij} = \sqrt{(F_i-F_j)^T(F_i-F_j)}^{-1}$, where $F_i$ is the vector of functionalities represented by population, jobs, percent low income, percent minority, percent adults, number of schools, shops, restaurants, etc.

\subsubsection{LSTM for Temporal Encoding}
\noindent
The LSTM network is chosen for temporal encoding. In short, the LSTM layers take the spatially encoded tensors for $l$ previous time periods as inputs and propagate them down the layers through a series of input, forget, cell state (memory), and output gates. To make the prediction, the encoded output from the last time step $h_{t-1}$ is decoded by fully connected layers to get the contribution (weights) of the time series on the final prediction. The LSTM structure is well-documented in a lot of papers \cite{Rodrigues2020,Chen2020}, and will not be discussed in detail here.



\subsection{Specific Examples: Gaussian Distributions and Mean-Variance Estimation}
\noindent 
As an important case, the Gaussian distribution with homoskedastic variance (HomoG-GCN and HomoG-GAT) represents the vast number of deterministic deep learning models that use mean squared errors as the training objective. With the homoskedastic Gaussian assumption, the dependent variable $y$ follows Gaussian distribution with mean $F_1(\mathbf{X}, w_1)$ and a constant variance $c$. The MLE with this homoskedastic Gaussian distribution is the same as the deterministic deep learning with the mean squared errors as the training objective because
\begin{flalign}
\log P(x;\mu,\sigma = c) 
                         &= \log \frac{1}{c \sqrt{2\pi}} - \frac{1}{2}\frac{(x-\mu)^2}{c^2} 
\end{flalign}
In other words, our benchmark example represents the predominant deterministic modeling technique in this field. 

The homoskedastic Gaussian can be extended to the heterskedastic Gaussian distribution, which resembles the mean-variance estimation (MVE), the most dominant method in uncertainty quantification literature. With the heteroskedastic Gaussian distribution, $y \sim \mathcal{N}(\theta) = \mathcal{N}(F_1(\mathbf{X}, w_1), F_2(\mathbf{X}, w_2)) = F_1(\mathbf{X}, w_1) + \mathcal{N}(0, F_2(\mathbf{X}, w_2))$. Essentially, the MVE uses two graph neural networks to capture the mean and variance separately, which is the same as the HetG-GCN and HetG-GAT models. Therefore, the two Gaussian examples in our probabilistic GNN framework can represent the two most common research methods, namely the deterministic spatiotemporal models and the MVE for uncertainty quantification. 

\subsection{Evaluation}
\noindent 
Performance metrics can be grouped into three categories: composite measures, point prediction quality, and uncertainty prediction quality. Table \ref{tab:eval} summarizes the evaluation metrics. The NLL loss is a composite metric to evaluate the joint quality of point and uncertainty estimates. The standard mean absolute error (MAE) and mean absolute percent error (MAPE) are used to evaluate point predictions.

\begin{table}[ht!]
    \centering
    \caption{Three Categories of Evaluation Metrics}
    \resizebox{\linewidth}{!}{
    \begin{tabular}{p{1.5cm}|c|c}
        \toprule
        \textbf{Category} & \textbf{Metric} & \textbf{Formula} \\
        \toprule
        Composite & Negative Log Likelihood & $NLL = -\sum_{i} \text{log } P_{W}(y_{i}|\mathbf{X_i})$ \\
        \midrule
        \multirow{2}{*}{Point} & Mean Absolute Error & $MAE=\frac{1}{N}\sum_{i} |y_i-\hat{y_i}|$ \\
        & Mean Absolute Percent Error & $MAPE = MAE / \Bar{y}$ \\
        \midrule
        \multirow{3}{*}{Uncertainty} & Calibration Error & $CE= \sum_{p=0}^{1}{|q(p)-p|}$\\
        & Mean PI Width & $MPIW = \frac{1}{n}{\sum_{i=1}^{N}{(U_i - L_i)}}$ \\
        & PI Coverage Probability & $PICP = \frac{1}{N}{\sum_{i=1}^{n}{\mathds{1}\{L_i \leq y_i \leq U_i \}}}$\\
        \bottomrule
    \end{tabular}}
    \label{tab:eval}
\end{table}

The quality of the uncertainty estimates is less straightforward to evaluate, because the ground truth distributions are unknown. Therefore, we design the calibration error (CE) metric and also visualize quantile-quantile plots to measure the distributional fit. Let $F_i(y)$ represent the cumulative distribution function of observation $i$. Define $q(p) = \mathbb{P}(F_i(y_i) \leq p)$ as the proportion of observations that actually fall into the $p$-th quantile of the estimated distribution. A well-calibrated model should generate distributions that align with the empirical distribution so that $q(p)=p$. Therefore, the calibration error associated with the quantile-quantile plots is defined as the sum of the deviation between the empirical quantiles and the predicted quantiles, approximated by a number of discrete bins from $[0,1]$: $CE= \sum_{p=0}^{1}{|q(p)-p|}$. 
An alternative metric is the simultaneous use of the Prediction Interval Coverage Probability (PICP) and the Mean Prediction Interval Width (MPIW) \cite{Pearce2018, Khosravi2011, Kabir2018}. Formally, $PICP = \frac{1}{n}{\sum_{i=1}^{n}{c_i}}$, where $c_i= \mathds{1}\{L_i \leq y_i \leq U_i \}$, that is an indicator variable of whether observation $i$ falls within the prediction interval bounded by $L_i$ and $U_i$. $MPIW = \frac{1}{n}{\sum_{i=1}^{n}{(U_i - L_i)}}$ measures the average width of the intervals. With significance level $1-\alpha$, the lower bound of the prediction interval is given by $L = F_i^{-1}(\frac{\alpha}{2})$, and the upper bound of the prediction interval by $U = F_i^{-1}(1-\frac{\alpha}{2})$, where $F_i^{-1}$ is the predicted inverse cumulative function of observation $i$. Using this approach, the evaluation of uncertainty quantification involves a tradeoff between PICP and MPIW. A high-quality prediction interval should be narrow while covering a large proportion of data points; however, wider MPIW naturally leads to larger PICP. Therefore, this tradeoff poses a challenge in model selection since it is difficult for one model to dominate in both metrics.

\section{Case Study}

Two case studies were conducted to examine the effect of deterministic and probabilistic assumptions on predicting travel demand, with a focus on estimating uncertainty. The case studies use data from the Chicago Transit Authority's (CTA) rail system and ridesharing in Chicago. 
This section describes data sources, the spatial and temporal characteristics of the two datasets, and the experiment setup. Our experiments are implemented in PyTorch and the source code is available at \url{https://github.com/sunnyqywang/uncertainty}. 

\subsection{Data}

The CTA rail and bus data was obtained through collaboration with the CTA, while the ridesharing data was sourced from the City of Chicago Open Data Portal\footnote{\url{https://data.cityofchicago.org/Transportation/Transportation-Network-Providers-Trips/m6dm-c72p}}.
The CTA rail system has tap-in records for each trip, which were aggregated into 15-minute intervals for temporal analysis. The system comprises 141 stations, arranged in a graph structure for spatial analysis. The spatial relationships can be identified by adjacency matrices constructed from Euclidean, connectivity, network and functional similarity between stations. Ridesharing trips are available at the census tract level, in 15-minute intervals. The graph structure is determined by the relationships between census tracts. As there is no explicit network, the ridesharing adjacency matrices are defined by Euclidean, connectivity (neighboring census tracts are considered connected), functional similarity only. Most census tracts have close to none ridesharing trips in most 15-minute intervals. Since learning sparse graphs are a topic of its own \cite{ye2021sparse}, the ridesharing dataset is filtered to include only those census tracts that, on average, had more than 30 trips per hour pre-COVID. This resulted in a total of 59 census tracts being used in the analysis.

\begin{figure}[ht!]
    \centering
    \subfloat[CTA Rail]{\includegraphics[height=0.4\linewidth]{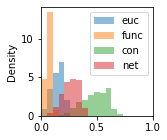}}
    \subfloat[Ridesharing]{\includegraphics[height=0.4\linewidth]{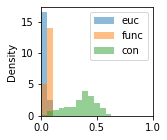}}
    \caption{Spatial Autocorrelation (Moran's I)}
    \label{fig:moran}
\end{figure}

Moran's I was calculated for each 15min ridership snapshot using each of the weight matrices to demonstrate that spatial autocorrelation exists through the defined adjacency matrices. Figure \ref{fig:moran} shows the histograms of Moran's I for both data sources. Most statistics have averages well above 0, indicating the existence of spatial autocorrelations defined by the adjacency matrices. Ridesharing trips appear to be less correlated with Euclidean distance and functional similarity, but the correlation between bordering tracts is very strong.

Five different explanatory variables are obtained for each data source. First, the system's own history, both recent (in the past 1.5 hours) and historical (last week same time period) is used to account for the temporal correlation. Second, the history of the other travel modes is included to mine the correlation between different modes. In the case of CTA rail, ridesharing and bus counts are included; and in the case of ridesharing, CTA rail and bus counts are included. Third, demographics\footnote{\url{https://www.census.gov/programs-surveys/acs/data.html}} and POIs \footnote{\url{https://planet.openstreetmap.org/}} are spatial covariates used in the calculation of functional similarity. Fourth, weather\footnote{\url{https://www.ncdc.noaa.gov/cdo-web/}} is a temporal covariate that is assumed to be the same in the whole study area. Lastly, the frequency of services in each spatial unit during each time period is used to indicate supply levels. In the case of ridesharing, we do not have supply information. Instead, we use the bus frequency as a proxy. Bus schedules are slower to respond during normal times, but after March 2020 the CTA is actively adjusting service to account for labor shortages and changing demand.

\subsection{Experiment Setup} \label{sec:exp_setup}

\begin{figure}[ht!]
    \centering
    \includegraphics[width=\linewidth]{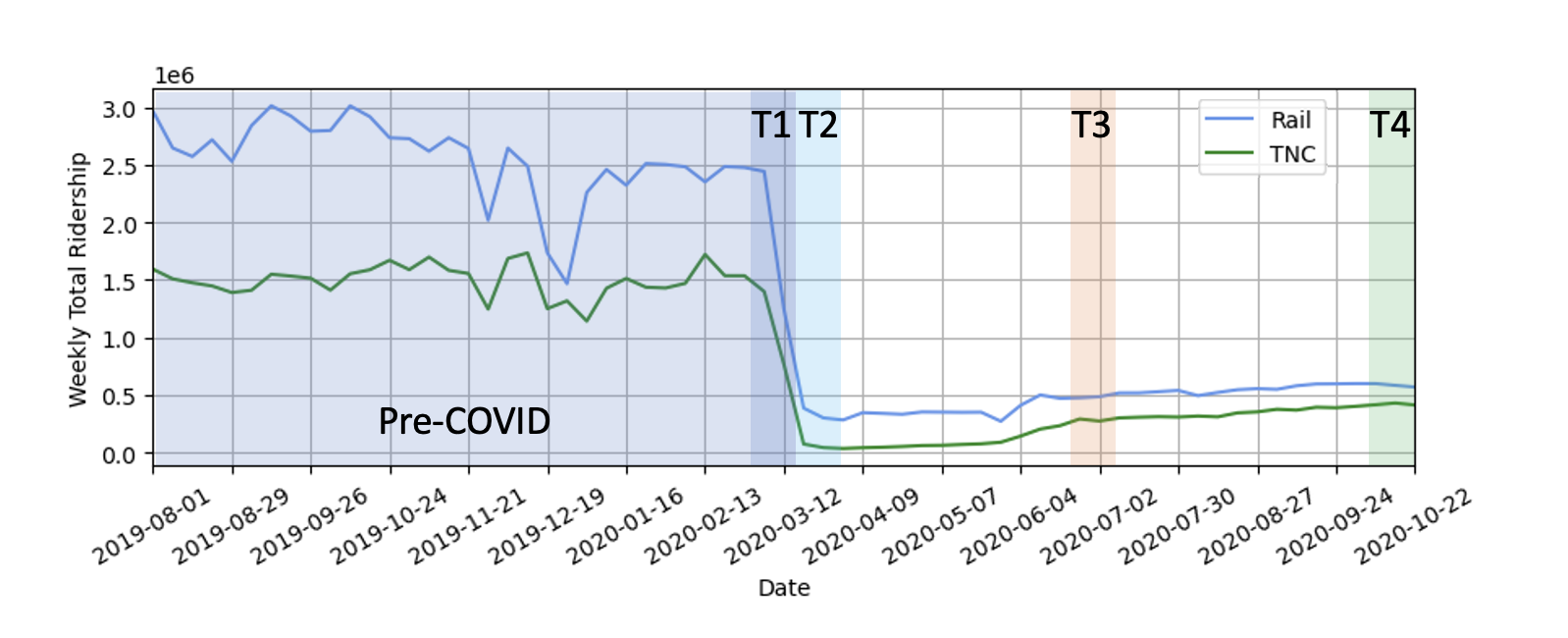}
    \caption{Average Daily Ridership (T1: Immediately Before; T2: Stay-at-home; T3: Initial Recovery; T4: Steady Recovery)}
    \label{fig:ridership_viz}
\end{figure}

The datasets were split into three subsets along the temporal dimension: train, validation, and test. The training set consists of data from August 1, 2019 to Feb 16, 2020, the validation set from Feb 17, 2020 to March 1, 2020, and the testing set consists of four different two-week periods during the course of COVID-19 pandemic in 2020: immediately before (March 2 to March 15), stay-at-home (March 16 to March 29), initial recovery (June 22 to July 5), and steady recovery (Oct 18 to Oct 31). Significant changes emerged starting March 2020 when the confirmed COVID-19 cases were growing rapidly and the city issued stay-at-home orders. Since then, the ridership has seen different stages of recovery. The changes in ridership of a few stations/census tracts are shown in Figure \ref{fig:ridership_viz}. Meanwhile, changes induced by COVID-19 provide an opportunity to test the temporal generalizability from pre- to post-COVID periods. 

\begin{figure}[ht!]
    \centering
    \resizebox{\linewidth}{!}{
    \includegraphics[width=\linewidth]{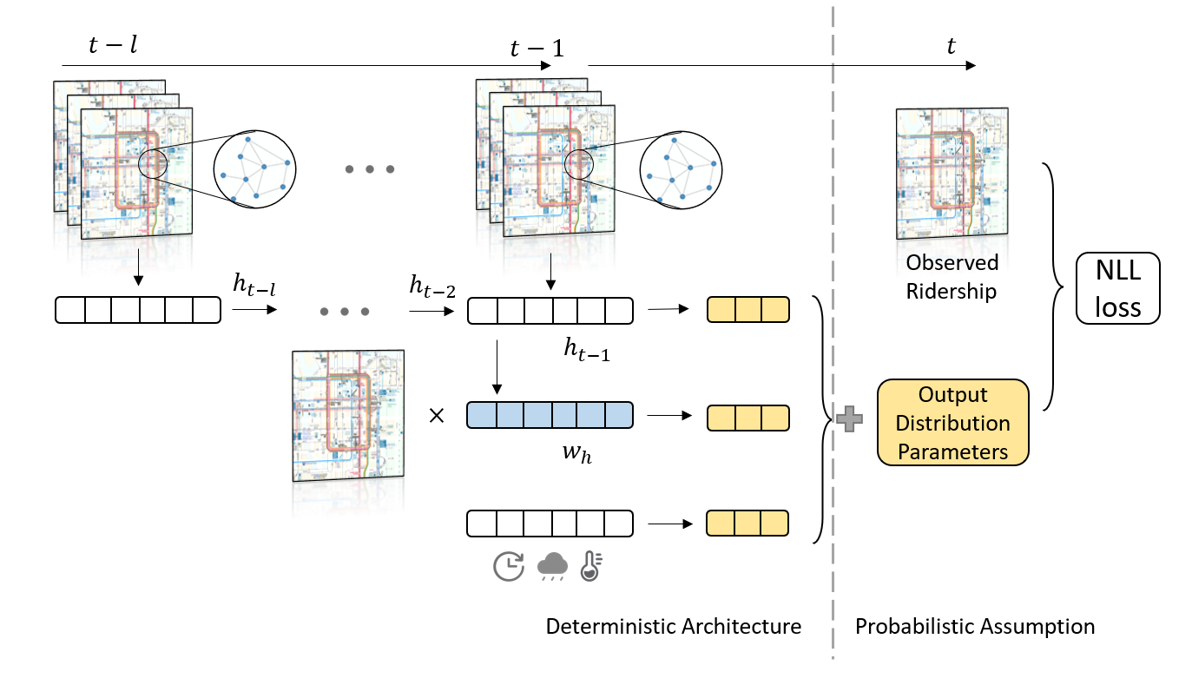}}
    \caption{Proposed model architecture. The model processes past demand information using spatial and temporal layers, with the addition of auxiliary information to produce  estimates of distributions parameters}
    \label{fig:arch}
\end{figure}

Several other architectural consideration have been taken into account besides the GCN/GAT spatial convolutions. Figure \ref{fig:arch} summarizes the deterministic architecture, consisting of three components: spatiotemporal layers (GCN/GAT+LSTM) for \textbf{r}ecent observed demand, and linear layer to connect last week's observation (\textbf{h}istory), and weather (tem\textbf{p}erature and \textbf{p}recipitation). The three components are indexed by $r,h,p$, respectively, and summed to the final prediction. 
First, due to the cyclic nature of travel demand, the travel demand time series can be decomposed into a weekly component ($\hat{\mathcal{Y}}^h_t$, $\hat{\sigma}^h_t$), and a time-of-day component ($\hat{\mathcal{Y}}^r_t$, $\hat{\sigma}^r_t$). 
The weekly component is calculated from a reference demand. A good reference demand is the observed value for the same region and time in the previous week. The weights $w_h$ are obtained from the LSTM encodings. Additionally, the recent demand used for LSTM network inputs are the deviations from last week's demand, and the decoded outputs are used to produce both the weekly and the time-of-day components. Intuitively, if the recent residual demand is very different from the reference, the reference should have a smaller weight on the final prediction, and the final prediction is more uncertain. The time-of-day component is directly obtained from LSTM encodings of the recent residual demand.
Next, weather is another source of temporal variation because extreme weather could have an evident impact on transit ridership. The model takes daily deviations from average of precipitation $P_T$ and temperature $T_E$, and multiply them with spatiotemporal weights $W^p, W^t \in \mathbb{R}^{2 \times T \times S}$ to get $[\hat{\mathcal{Y}}_t^p, \hat{\sigma}_t^p]$.


For each dataset and each deterministic architecture, six probabilistic assumptions are tested: homoskedastic Gaussian (HomoG), Poisson (Pois), heteroskedastic Gaussian (HetG), truncated Gaussian (TG), Laplace (Lap), ensembled Gaussian (GEns). In the HomoG model, a search for the best standard deviation was done. Values are searched in multiples of $\Bar{y}$: $\frac{1}{4}, \frac{1}{2}, \frac{3}{4}, 1$, and the best value for the variance is $\frac{1}{2}\Bar{y}$ for both datasets. In GEns, five top HetG models by validation set loss are selected to create an ensemble model. The ensemble distribution is given by \cite{Lakshminarayanan2016}: $\mathcal{Y}_* \sim \mathcal{N}(\frac{1}{K} \sum_k{\mathcal{Y}_k}, \frac{1}{K} \sum_k{(\sigma_k^2 + \mathcal{Y}_k^2)} - \mathcal{Y}_*^2 )$. 





\section{Results and Discussion}
We present and compare the model performances of the Prob-GNNs under six probabilistic and two deterministic assumptions on two separate datasets. First, we analyze the model performance for periods immediately after the training period, that is, from March 2 to March 15, 2020. We demonstrate that probabilistic assumptions can influence the model performance more significantly than deterministic architectures. Next, we show that uncertainty predictions are important especially when point predictions become unreliable during system disruptions, by applying models trained with the pre-COVID training set to the post-COVID test sets. Lastly, we discuss the spatiotemporal patterns of uncertainty revealed by the models.

\begin{figure}[ht!]
    \centering
    \includegraphics[width=\linewidth]{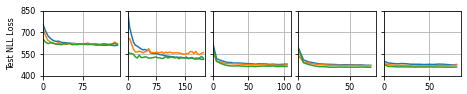}\\
    \includegraphics[width=\linewidth]{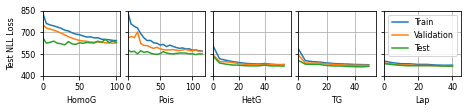}
    \caption{Training Curves. Top: GCN; Bottom: GAT}
    \label{fig:training_curves}
\end{figure}

First, to illustrate the models' proper convergence, Figure \ref{fig:training_curves} presents the GCN model training curves for all probabilistic assumptions trained on CTA Rail. All models demonstrated successful convergence under the same learning rate. However, the number of epochs required for convergence varied among the models. The two-parameter distributions (HetG, TG, Laplace) converged faster, and achieved a lower NLL compared to the one-parameter distributions (HomoG, Pois).

Model performances of CTA Rail and ridesharing data, on the test set between March 2 and March 15, 2020, are tabulated in Tables \ref{tab:model_perf}. The table has two panels for GCN and GAT models, respectively. Each panel presents the six different probabilistic assumptions presented in Section \ref{sec:exp_setup}. Model performances are measured on three categories of metrics: composite (NLL), uncertainty (calibration error, MPIW, and PICP), and point (MAE, MAPE). 
In the tables, the intended level of coverage was set to 95\%. 
The next two sections discuss the findings from Table \ref{tab:model_perf} in detail.

\begin{table*}[ht!]
    \centering
    \caption{Model Performance (Test Period: Immediately Before - March 2 to March 15, 2020)}
    \resizebox{\linewidth}{!}{
    \begin{tabular}{p{2cm}|P{1.2cm}|P{1.2cm}|P{1.2cm}|P{1.2cm}|P{1.2cm}|P{1.2cm}||P{1.2cm}|P{1.2cm}|P{1.2cm}|P{1.2cm}|P{1.2cm}|P{1.2cm}}
        \toprule
        & \multicolumn{6}{c||}{CTA Rail} & \multicolumn{6}{c}{Ridesharing}\\
        \midrule
        & Comp & \multicolumn{3}{c|}{Uncertainty Prediction} & \multicolumn{2}{c||}{Point Prediction} & Comp & \multicolumn{3}{c|}{ Uncertainty Prediction} & \multicolumn{2}{c}{Point Prediction}\\
        \midrule
        Model & NLL & Cal. Err & MPIW & PICP & MAE & MAPE & NLL & Cal. Err & MPIW & PICP & MAE & MAPE\\
        \midrule
        HomoG-GCN & 612.5 & 0.146 & 64.53 & 97.9\% & 7.64 & 18.5\% & 225.6 & 0.114 & 42.63 & 97.7\% & 5.81 & 25.2\% \\
        Pois-GCN & 529.2 & 0.075 & 18.04 & 84.9\% & \textbf{7.18} & \textbf{17.4\%} & 211.9 & 0.083 & 15.53 & 82.7\% & 5.81 & 25.2\% \\
        HetG-GCN & 462.9 & 0.049 & 39.04 & 96.5\% & 7.67 & 18.6\% & 187.6 & 0.019 & 27.32 & 93.3\% & 5.80 & 25.1\% \\
        TG-GCN & 480.7 & \textbf{0.021} & 39.37 & 95.6\% & 7.61 & 18.5\% & 196.7 & 0.029 & \textbf{30.63} & \textbf{95.2\%} & 6.06 & 26.3\% \\
        Lap-GCN & 460.2 & 0.026 & 43.48 & 97.0\% & 7.52 & 18.2\% & 187.1 & 0.022 & 34.12 & 96.5\% & 5.81 & 25.2\% \\
        GEns-GCN & \textbf{459.6} & 0.036 & \textbf{37.48} & \textbf{95.7\%} & 7.42 & 18.0\% & \textbf{185.6} & \textbf{0.017} & 28.01 & 93.9\% & \textbf{5.70} & \textbf{24.7\%} \\
        \midrule
        \midrule
        HomoG-GAT & 620.4 & 0.141 & 64.54 & 97.8\% & 8.13 & 19.7\% & 230.7 & 0.112 & 43.26 & 96.9\% & 6.52 & 28.2\% \\
        Pois-GAT & 572.2 & 0.100 & 18.11 & 81.8\% & 7.98 & 19.4\% & 214.9 & 0.093 & 15.62 & 82.0\% & \textbf{6.06} & \textbf{26.3\%} \\
        HetG-GAT & 472.3 & 0.058 & \textbf{39.58} & \textbf{96.3\%} & 7.84 & 19.0\% & 188.3 & 0.041 & \textbf{31.43} & \textbf{95.7\%} & 6.07 & 26.3\% \\
        TG-GAT & 481.6 & \textbf{0.026} & 40.96 & 96.6\% & 7.69 & 18.7\% & 199.9 & 0.024 & 32.54 & 95.7\% & 6.55 & 28.4\% \\
        Lap-GAT & \textbf{464.3} & 0.037 & 42.77 & 97.7\% & 7.63 & 18.5\% & 189.5 & \textbf{0.021} & 36.08 & 96.5\% & 6.16 & 26.7\% \\
        GEns-GAT & 470.5 & 0.037 & 42.49 & 96.8\% & \textbf{7.52} & \textbf{18.3\%}  & \textbf{188.1} & 0.044 & 32.28 & 96.0\% & 6.18 & 26.8\% \\
        \bottomrule
    \end{tabular}
    }
    \label{tab:model_perf}
\end{table*}

\subsection{Significance of Probabilistic and Deterministic Assumptions}\label{sec:prob}

The quality of uncertainty quantification varies remarkably with probabilistic assumptions. The performance is significantly higher in the probabilistic models with two parameters (HetG and Lap) compared to the single-parameter models (HomoG and Pois). Among all GCN models, the average NLL and calibration error for two-parameter distributions in the CTA Rail data is $465.9$ and $0.033$, and for one-parameter distributions $570.8 (+22.5\%)$ and $0.11 (3.3 \times)$. Similar observations can be made in the ridesharing dataset, where the average NLL is $218.8$ vs. $189.3 (+15.4\%)$ and the calibration error is $0.017$ vs. $0.099 (5.8 \times)$. Similar observations can be made for GAT models.

\begin{figure}[ht!]
    \centering
    \subfloat[]{\includegraphics[width=0.5\linewidth]{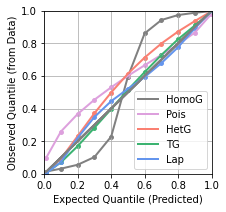}}
    \subfloat[]{\includegraphics[width=0.5\linewidth]{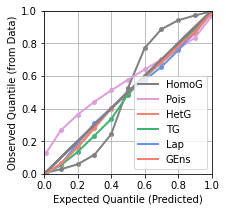}}
    \caption{Calibration Plot of GCN Models (a) CTA rail and (b) Ridesharing. The line $y=x$ represents a perfectly-calibrated model. The calibration error (CE) corresponds to the areas between the $y=x$ line and the empirical quantiles.}
    \label{fig:qq_calibration}
\end{figure}

Truncated Gaussian and Laplace distribution have the overall best distributional fit. Figure \ref{fig:qq_calibration} illustrates the distributional fit by plotting the empirical quantiles $q(p)$ against the theoretical quantiles $p$ for both datasets across all the GCN models. The line $y=x$ represents a perfectly-calibrated model. The calibration error in Table \ref{tab:model_perf} corresponds to the areas between the $y=x$ line and the empirical quantiles. Poisson and homoskedastic Gaussian are very far away from the $y=x$ line, while truncated Gaussian and Laplace trace the line most closely. Although the Heteroskedastic Gaussian works the best with GCN models on ridesharing data, the truncated Gaussian and Laplace distribution had more stable performances across the datasets and models, and the curves are closer to the $y=x$ line. 

In contrast to the strong influence probabilistic assumptions have on the quality of uncertainty predictions, little influence is exerted on the point prediction quality from the probabilistic assumptions. For both datasets and both GCN and GAT, between different probabilistic assumptions, the performance gap between the best and the worst point estimate is around 4\%, significantly less than that of the composite and the uncertainty metrics. 

The variations in predictive performance caused by the deterministic assumptions are also small compared to the probabilistic assumptions. Comparing the GCNs and GATs, the error metrics are similar across the models, and the performance patterns across probabilistic assumptions remain the same for GAT models. The one-parameter probabilistic assumptions are likely to be affected more by deterministic architectures. For all probabilistic assumptions except for Poisson, less than a 3\% difference in NLL loss is observed between GCN and GAT models for the same dataset. 
Although the GAT allows the model to learn spatial relationships and is hence more flexible, the predefined adjacency matrices serve as domain knowledge to reduce the learning complexity. In theory, the GAT setup should perform better with more complex relationships or with larger sample sizes, and the GCN setup is better for more efficient learning under limited sample sizes. In this case, there is not a large difference between the two architectures.

\subsection{Implications of Probabilistic Assumptions}

The probabilistic assumptions not only suggest the distributional fit, but also have practical implications on the ridership patterns. We then discuss these implications in detail below.

\subsubsection{Heteroskedasticity vs. Homoskedasticity (HetG vs. HomoG)} HetG forces a constant variance across all observations, resulting in an inaccurate representation of the data. The HomoG models achieve around $20\%$ higher NLL loss and $1.5$ to $2$ times higher calibration error than the HetG models. For example, the NLL loss of HetG-GCN is 462.9, 24\% lower than the NLL loss (612.5) of HomoG-GCN for CTA Rail. Although predicting the same variance is inaccurate distributionally, the average MPIW and PICP are not much worse compared to their heteroskedastic counterparts since the variance scale is searched and fixed. Additionally, the inaccurate distributional assumption does not affect the quality of the point estimate, and point estimate errors from HomoG models are only 1-2\% higher than the best model, as the likelihood improvement can only come from a more accurate prediction of the mean.
Regardless, having one fixed uncertainty parameter not only is an inaccurate representation, but also limits the model's flexibility to adapt to sharp changes in ridership magnitude.

\subsubsection{Continuous vs. Discrete (HomoG vs. Pois)} Both HomoG and Pois have the worst NLL loss among all probabilistic assumptions tested. Despite its discreteness, the Poisson assumption yields similar or worse NLL and calibration error than HomoG since it forces equal mean and variance. However, for both datasets, the prediction intervals significantly under-predict the demand, meaning that the variance should be larger than the mean. The PICP of the Poisson prediction interval indicates the magnitude of variance compared to the mean. The smaller the PICP, the larger the variance. 

\subsubsection{Real-line vs. non-negative support (HetG vs. TG)} Since the time resolution is 15min, left-truncation at 0 significantly improves the model performance as many predictions have means close to 0. TG is the best-performing model for the CTA Rail and the second-best for ridesharing. Even though the average ridership is around 40 per station and 23 per census tract, significant heterogeneity exists among different stations/census tracts. Sparsity (zero ridership) is an issue to be considered in short-term or real-time demand predictions.

\subsubsection{Gaussian vs. Exponential tails (HetG vs. Lap)} Both distributions are characterized by two parameters and can be trained to describe the data accurately. The weights of the tails are different, and the model performances on the two probabilistic assumptions suggest behavioral differences. Compared to HetG, Lap has a heavier tail; hence the prediction intervals tend to be larger, covering more observations in more extreme cases. In all cases, Lap covers more points than intended.  

\subsubsection{Single vs. ensembled models (HetG vs. GEns)} Ensembling only makes marginal improvements to the model results. The NLL loss, calibration error, and MAE improve by less than 1\% in all cases. For example, the biggest improvement occurs in the GAT models for CTA Rail data, with NLL loss for HetG-GAT being 462.9 and for GEns-GAT 459.6. 
Since ensembling aims at reducing model uncertainty, its ineffectiveness suggests that different training instances of the neural networks produce similar results, and the model uncertainty is low. 

We also constructed prediction intervals with Monte Carlo dropouts to further illustrate that model uncertainty is much smaller than the data uncertainty inherent in the data generation process. Monte Carlo dropout approximates Bayesian neural networks and measures the model uncertainty by applying dropout at test time, treating each as an instance from the space of all available models. However, since the different training instances produce similar results, Monte Carlo dropout fails to capture the full picture. Its prediction intervals were very narrow with PICPs between 30\% - 40\% for both datasets. Additionally, since we applied dropout at test time, the point prediction loses the benefit of dropout regularization and performs worse than other models.

\subsection{Model Performance under System Disruption} \label{sec:disrupt}

The drastic change in pre- and post-COVID periods presents a unique opportunity to test the generalizability of Prob-GNNs. The previous sections use the periods immediately following the training set for validation and testing. This section compares the predictive performance at different stages of the COVID-19 pandemic, by applying the models trained with the pre-COVID training set to three post-COVID time periods. Table \ref{tab:model_perf_test} presents the results for CTA Rail and ridesharing. Since HomoG and Pois have poor performance in the previous test set, they are excluded from this comparison.

\begin{table*}[ht!]
    \centering
    \caption{Model Generalization during Disruptions}
    \resizebox{\linewidth}{!}{
    \begin{tabular}{p{2cm}|P{1.2cm}|P{1.2cm}|P{1.2cm}|P{1.2cm}|P{1.2cm}|P{1.2cm}||P{1.2cm}|P{1.2cm}|P{1.2cm}|P{1.2cm}|P{1.2cm}|P{1.2cm}}
        \toprule
        & \multicolumn{6}{c||}{CTA Rail} & \multicolumn{6}{c}{Ridesharing}\\
        \midrule
        & Comp & \multicolumn{3}{c|}{Uncertainty Prediction} & \multicolumn{2}{c||}{Point Prediction} & Comp & \multicolumn{3}{c|}{ Uncertainty Prediction} & \multicolumn{2}{c}{Point Prediction}\\
        \midrule
        Model & NLL & Cal. Err & MPIW & PICP & MAE & MAPE & NLL & Cal. Err & MPIW & PICP & MAE & MAPE\\
        \midrule
        \multicolumn{13}{c}{Stay-at-home (March 16 - March 29, 2020)}\\
        \midrule
        HetG-GCN & 397.9 & 0.258 & 16.44 & 96.0\% & 4.92 & 105\% & 165.2 & 0.291 & 11.89 & 88.1\% & 4.40 & 198\% \\
        TG-GCN & \textbf{388.9} & \textbf{0.176} & 16.54 & 95.2\% & 4.38 & 93.6\%  & 163.4 & \textbf{0.114} & 13.04 & 97.1\% & \textbf{2.96} & \textbf{133\%} \\
        Lap-GCN & 405.8 & 0.247 & 17.65 & 98.7\% & \textbf{4.34} & \textbf{92.9\%} & \textbf{148.3} & 0.217 & \textbf{13.41} & \textbf{96.7\%} & 3.70 & 167\% \\
        GEns-GCN & 400.4 & 0.263 & \textbf{16.14} & \textbf{95.6\%} & 4.89 & 105\% & 162.4 & 0.286 & 11.94 & 89.9\% & 4.20 & 189\% \\
        \midrule
        \midrule
        \multicolumn{13}{c}{Initial Recovery (June 22 - July 5, 2020)}\\
        \midrule
        HetG-GCN & 413.7 & 0.214 & \textbf{18.19} & \textbf{94.9\%} & 4.46 & 61.6\% & 161.9 & 0.171 & 12.39 & 90.2\% & 3.57 & 68.3\% \\
        TG-GCN & 436.2 & \textbf{0.174} & 18.29 & 94.3\% & 4.37 & 60.3\% & 173.3 & \textbf{0.037} & 12.91 & 93.8\% & 2.88 & 55.1\% \\
        Lap-GCN & \textbf{409.4} & 0.190 & 18.83 & 98.3\% & \textbf{3.95} & \textbf{54.5\%} & 161.7 & 0.071 & \textbf{13.81} & \textbf{94.2\%} & \textbf{2.82} & \textbf{54.0\%} \\
        GEns-GCN & 412.4 & 0.213 & 18.45 & 95.0\% & 4.4 & 60.7\% & \textbf{152.2} & 0.112 & 11.22 & 92.0\% & 2.88 & 55.1\% \\
        \midrule
        \midrule
        \multicolumn{13}{c}{Steady Recovery (Oct 12 - Oct 25, 2020)}\\
        \midrule
        HetG-GCN & \textbf{387.6} & \textbf{0.017} & 16.79 & 93.7\% & \textbf{3.38} & \textbf{36.2\%} & 174.7 & 0.128 & 15.80 & 89.1\% & 4.33 & 51.4\% \\
        TG-GCN & 389.5 & 0.030 & \textbf{18.46} & \textbf{95.9\%} & 3.90 & 41.7\% & 186.0 & \textbf{0.035} & 17.16 & 92.9\% & 4.03 & 47.8\% \\
        Lap-GCN & 406.6 & 0.057 & 21.63 & 96.1\% & 4.12 & 44.1\% & \textbf{170.6} & 0.073 & \textbf{18.62} & \textbf{94.4\%} & \textbf{3.93} & \textbf{46.7\%} \\
        GEns-GCN & 391.1 & 0.053 & 20.28 & 95.1\% & 3.44 & 36.8\% & 172.7 & 0.098 & 14.75 & 89.2\% & 3.95 & 47.0\% \\
        \midrule
    \end{tabular}
    }
    \label{tab:model_perf_test}
\end{table*}

All the error metrics increase under the significant domain shifts from pre- to post-COVID. In the stay-at-home period, the average ridership for both systems dropped to less than 10\% of pre-COVID levels. The NLL, MPIW, and MAE are not comparable to pre-COVID levels because they are influenced by the magnitude of ridership, while calibration error, PICP, and MAPE are unit-free and can be compared to values in Table \ref{tab:model_perf}. Unsurprisingly, the performance during the stay-at-home period is relatively poor but slowly rebounds with the recovery of the ridership. 

In the case where significant disruptions happen in the system, the point predictions fail miserably, but the uncertainty predictions stay accurate and indicative of the changing situation.
The MAPE for the test set in Table \ref{tab:model_perf} was 18\% for CTA and 25\% for ridesharing. For the three additional periods, the MAPEs are 93\%, 43\%, 36\% for CTA and 133\%, 54\%, and 47\% for ridesharing, respectively. The uncertainty predictions recovered a lot faster than MAPE. The calibration error returned to pre-COVID levels at the steady recovery stage, although the point prediction error is still 10\% higher than before. If only a 95\% prediction interval is considered, even at the stay-at-home home stage, we can achieve pretty precise prediction intervals (95.6\% and 96.7\% for CTA and ridesharing). 

The generalizability between different two-parameter distributions is similar, although each has distinct characteristics. TG-GCN excels at low ridership; Lap-GCN is heavy-tailed and more conservative; while HetG-GCN relies on the similarity between traning and testing domains. Since the situation is evolving constantly, there is no probabilistic assumption that dominates all scenarios. But some general conclusions can be drawn. In most cases, TG-GCN has the best calibration error due to its left-truncation. Enforcing non-negativity is beneficial since the ridership haven't recovered to pre-COVID levels for either CTA or ridesharing. Lap-GCN typically produces the best NLL and point prediction due to its more distributed shape, reducing overfitting to pre-COVID data. As ridesharing trips become more spontaneous in post-COVID times, Lap-GCN's conservative prediction intervals outperform others.



\subsection{Spatiotemporal Uncertainty}

\begin{figure*}[ht!]
    \centering
    \includegraphics[width=0.23\linewidth]{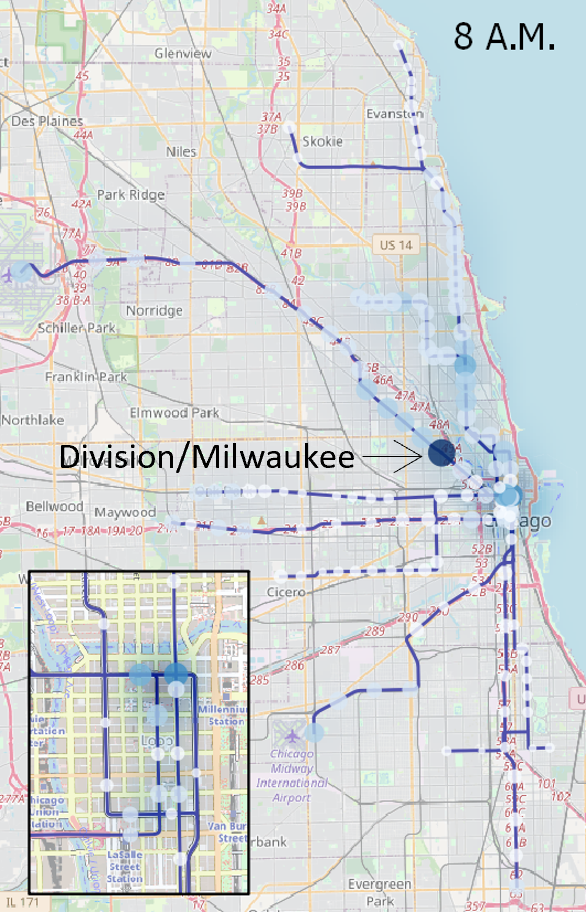}\hspace{1mm}
    \includegraphics[width=0.23\linewidth]{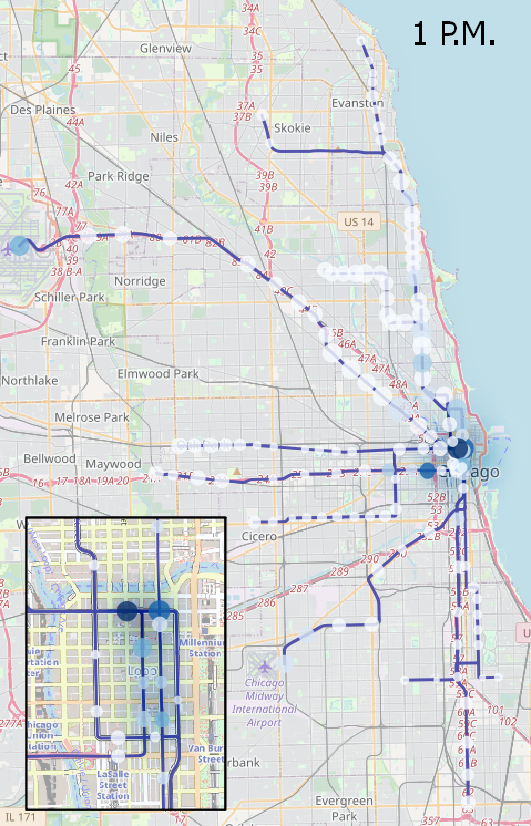}\hspace{1mm}
    \includegraphics[width=0.23\linewidth]{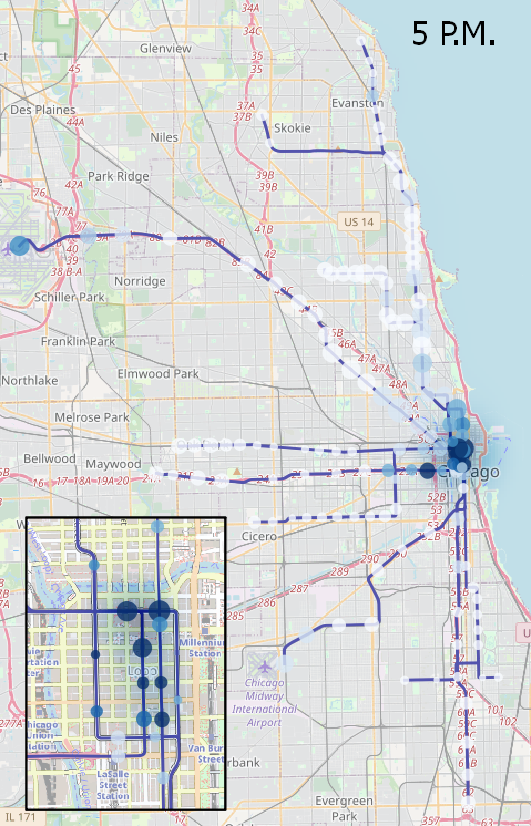}\hspace{1mm}
    \includegraphics[width=0.23\linewidth]{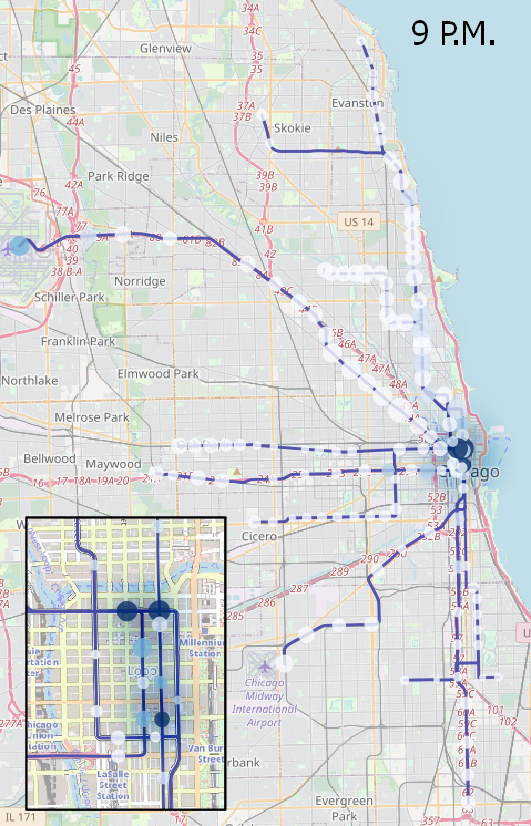}\\
    \includegraphics[width=\linewidth]{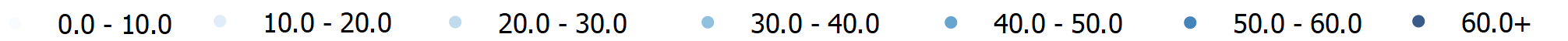}
    \caption{Spatiotemporal Uncertainty. Standard deviations of estimated CTA Rail station tap-ins in the 15min periods starting at 8 A.M. (morning peak), 1 P.M. (midday), 5 P.M. (afternoon peak), and 9 P.M. (evening).}
    \label{fig:st_unc}
\end{figure*}

Travel demand uncertainty has both spatial and temporal patterns: spatially, uncertainty is higher at stations with higher volumes, and temporally, uncertainty is higher in the afternoons. Figure \ref{fig:st_unc} shows the spatial distribution of predicted uncertainty for CTA Rail at different times of the day during the steady recovery test period (Oct 12 - 25, 2020). The left bottom corner of each figure zooms in on the ``loop'', the downtown of Chicago. Uncertainty is proportional to both the size and the color of the circles, with darker and larger circles indicating higher uncertainty. 

Spatially, uncertainty is higher at busier stations. During the same time period, darker and larger circles appear near downtown, transfer stations, and airports. Statistically, if the occurrence of each potential trip is a random Bernoulli process with probability $p$, and we do not have further information on the values of $p$ for each trip, having more potential trips $n$ will yield higher uncertainty in the sum. The number of observed trips has a binomial distribution with variance $np(1-p)$, which is proportional to $n$. Practically, the trips from downtown, transfer stations, and airports are usually more diverse and complex in nature, which promotes spontaneity in trip-making, resulting in higher uncertainty.

Temporally, uncertainty is higher during the afternoon peak. The uncertainty during the morning peak is even generally smaller than that during midday and in the evening, although the morning peak has significantly higher ridership. Statistically, this observation could be attributed to having knowledge about some of the $p$'s. Since the morning peak primarily consists of commuting trips, the probability of the trips happening is higher than recreational trips, which tend to happen during midday and afternoons.

Spatiotemporal uncertainty predictions can inform strategic decision-making regarding capacity buffers. First, the framework could be used to identify bottlenecks and outliers in the system. For example, the station Division/Milwaukee on the blue line has unusually high uncertainty during the morning peak. Further investigation can be done to identify the reasons for the abnormal behavior and perform demand management. Moreover, different strategies are needed for different types of systems. In systems with a fixed, relatively large capacity, such as stations along the same subway line, uncertainty at the busiest stations at peak times is the most critical, as the rest of the line will have quite a lot of excess capacity. Therefore, understanding uncertainty in low-demand regions is important for behavioral analysis, but less critical for service planning. However, in systems that are built to meet demand, such as ride-hailing, uncertainty in lower-demand regions is as important as higher-demand regions, as the supply in those regions will be proportionally low. Re-balancing actions will be needed across the system.

\section{Conclusion}
Despite the importance of uncertainty in travel demand prediction, past studies use deep learning to predict only the average travel demand but not quantify its uncertainty. To address this gap, this study proposes a framework of Prob-GNNs to quantify the spatiotemporal uncertainty of travel demand. The framework is concretized by six probabilistic assumptions (HomoG, Pois, HetG, TG, GEns, Lap) and two deterministic ones (GCN and GAT), which are applied to transit and ridesharing data, yielding the following conclusions.

First, the Prob-GNN framework can successfully quantify spatiotemporal uncertainty in travel demand with empirical evidence demonstrated on the transit and ridesharing datasets. In both cases, the Prob-GNNs can accurately characterize the probabilistic distributions of travel demand, while retain the point predictions of a similar quality to the deterministic counterpart (HomoG). 
Second, the probabilistic assumptions have a much more substantial impact on model performance than the deterministic ones. The two deterministic architectures lead to less than a 3\% difference in NLL loss, while a wise choice of probabilistic assumptions could drastically improve model performance. Specifically, the two-parameter distributions (e.g., truncated Gaussian) achieve the highest predictive performance, which is 20\% higher in NLL and 3-5 times lower in calibration errors in comparison to the one-parameter baseline distributions.
Third, the Prob-GNNs enhance model generalizability under significant system disruptions. By applying the models trained on pre-COVID data to three post-COVID periods, we show that the point predictions fail to generalize but the uncertainty predictions remain accurate and successfully reflect the evolving situations. Even under significant domain shifts, the difference of predictive performance among the two-parameter distributions is minor. 
Lastly, Prob-GNNs can reveal spatiotemporal uncertainty patterns. Uncertainty is spatially concentrated on the stations with higher travel volume, and temporally concentrated on the afternoon peak hours. In addition, the Prob-GNNs can identify the stations with abnormally large uncertainty, which can inform real-time traffic controls to proactively address potential system disruptions. 


Future studies could advance this work by taking further theoretical or empirical efforts. Theoretically, Prob-GNN is a parametric uncertainty quantification method, which should be compared to Bayesian and non-parametric methods in terms of prediction quality. Empirically, as transportation patterns are diverse across modes and cities, it is important to test our framework under varying data scales and contexts to further corroborate the conclusions above to better inform policy-making. In addition, our framework can be generally applied to the prediction of origin-destination flows, travel safety, or even climate risks. Since uncertainty in urban systems has broad policy implications, future studies could integrate the Prob-GNNs with robust optimization methods to enhance urban resilience.

\section*{Acknowledgment}
This material is based upon work supported by the U.S. Department of Energy’s Office of Energy Efficiency and Renewable Energy (EERE) under the Vehicle Technology Program Award Number DE-EE0009211. The views expressed herein do not necessarily represent the views of the U.S. Department of Energy or the United States Government. The authors would also like to thank the Chicago Transit Authority for providing access to public transit ridership data. S.W. also acknowledges the partial funding support from the Research Opportunity Seed Fund (ROSF) 2023 at the University of Florida.

\bibliographystyle{IEEEtran}

\bibliography{references}

\begin{thebibliography}{10}
\providecommand{\url}[1]{#1}
\csname url@samestyle\endcsname
\providecommand{\newblock}{\relax}
\providecommand{\bibinfo}[2]{#2}
\providecommand{\BIBentrySTDinterwordspacing}{\spaceskip=0pt\relax}
\providecommand{\BIBentryALTinterwordstretchfactor}{4}
\providecommand{\BIBentryALTinterwordspacing}{\spaceskip=\fontdimen2\font plus
\BIBentryALTinterwordstretchfactor\fontdimen3\font minus
  \fontdimen4\font\relax}
\providecommand{\BIBforeignlanguage}[2]{{%
\expandafter\ifx\csname l@#1\endcsname\relax
\typeout{** WARNING: IEEEtran.bst: No hyphenation pattern has been}%
\typeout{** loaded for the language `#1'. Using the pattern for}%
\typeout{** the default language instead.}%
\else
\language=\csname l@#1\endcsname
\fi
#2}}
\providecommand{\BIBdecl}{\relax}
\BIBdecl

\bibitem{Yang2019}
\BIBentryALTinterwordspacing
K.~Yang, J.~Tu, and T.~Chen, ``{Homoscedasticity: An overlooked critical
  assumption for linear regression},'' \emph{General Psychiatry}, vol.~32,
  no.~5, p. 100148, oct 2019. [Online]. Available:
  \url{/pmc/articles/PMC6802968/ /pmc/articles/PMC6802968/?report=abstract
  https://www.ncbi.nlm.nih.gov/pmc/articles/PMC6802968/}
\BIBentrySTDinterwordspacing

\bibitem{Guo2021}
\BIBentryALTinterwordspacing
X.~Guo, N.~S. Caros, and J.~Zhao, ``{Robust matching-integrated vehicle
  rebalancing in ride-hailing system with uncertain demand},''
  \emph{Transportation Research Part B: Methodological}, vol. 150, pp.
  161--189, aug 2021. [Online]. Available:
  \url{https://linkinghub.elsevier.com/retrieve/pii/S0191261521001004}
\BIBentrySTDinterwordspacing

\bibitem{guo2022data}
X.~Guo, Q.~Wang, and J.~Zhao, ``Data-driven vehicle rebalancing with predictive
  prescriptions in the ride-hailing system,'' \emph{IEEE Open Journal of
  Intelligent Transportation Systems}, vol.~3, pp. 251--266, 2022.

\bibitem{Yao2018}
H.~Yao, F.~Wu, J.~Ke, X.~Tang, Y.~Jia, S.~Lu, P.~Gong, Z.~Li, J.~Ye, and
  D.~Chuxing, ``{Deep multi-view spatial-temporal network for taxi demand
  prediction},'' \emph{32nd AAAI Conference on Artificial Intelligence, AAAI
  2018}, pp. 2588--2595, 2018.

\bibitem{wu2020inductive}
\BIBentryALTinterwordspacing
Y.~Wu, D.~Zhuang, A.~Labbe, and L.~Sun, ``Inductive graph neural networks for
  spatiotemporal kriging,'' \emph{Proceedings of the AAAI Conference on
  Artificial Intelligence}, vol.~35, no.~5, pp. 4478--4485, May 2021. [Online].
  Available: \url{https://ojs.aaai.org/index.php/AAAI/article/view/16575}
\BIBentrySTDinterwordspacing

\bibitem{liu2022universal}
F.~Liu, J.~Wang, J.~Tian, D.~Zhuang, L.~Miranda-Moreno, and L.~Sun, ``A
  universal framework of spatiotemporal bias block for long-term traffic
  forecasting,'' \emph{IEEE Transactions on Intelligent Transportation
  Systems}, 2022.

\bibitem{zhuang2020compound}
D.~Zhuang, S.~Hao, D.-H. Lee, and J.~G. Jin, ``From compound word to
  metropolitan station: Semantic similarity analysis using smart card data,''
  \emph{Transportation Research Part C: Emerging Technologies}, vol. 114, pp.
  322--337, 2020.

\bibitem{Ke_2017}
J.~Ke, H.~Zheng, H.~Yang, and X.~Chen, ``Short-term forecasting of passenger
  demand under on-demand ride services: A spatio-temporal deep learning
  approach,'' \emph{Transportation Research Part C-emerging Technologies},
  2017.

\bibitem{Ye2020a}
J.~Ye, J.~Zhao, K.~Ye, and C.~Xu, ``{Multi-STGCnet: A Graph Convolution Based
  Spatial-Temporal Framework for Subway Passenger Flow Forecasting},'' in
  \emph{Proceedings of the International Joint Conference on Neural
  Networks}.\hskip 1em plus 0.5em minus 0.4em\relax Institute of Electrical and
  Electronics Engineers Inc., jul 2020.

\bibitem{Liu2020}
L.~Liu, J.~Chen, H.~Wu, J.~Zhen, G.~Li, and L.~Lin, ``{Physical-Virtual
  Collaboration Modeling for Intra- and Inter-Station Metro Ridership
  Prediction},'' \emph{IEEE Transactions on Intelligent Transportation
  Systems}, pp. 1--15, 2020.

\bibitem{wang2020_asu}
S.~Wang, B.~Mo, and J.~Zhao, ``Deep neural networks for choice analysis:
  Architecture design with alternative-specific utility functions,''
  \emph{Transportation Research Part C: Emerging Technologies}, vol. 112, pp.
  234--251, 2020.

\bibitem{wang2020_econ_info}
S.~Wang, Q.~Wang, and J.~Zhao, ``Deep neural networks for choice analysis:
  Extracting complete economic information for interpretation,''
  \emph{Transportation Research Part C: Emerging Technologies}, vol. 118, p.
  102701, 2020.

\bibitem{wang2020_mtldnn}
------, ``Multitask learning deep neural networks to combine revealed and
  stated preference data,'' \emph{Journal of Choice Modelling}, p. 100236,
  2020.

\bibitem{xiong2020dynamic}
X.~Xiong, K.~Ozbay, L.~Jin, and C.~Feng, ``Dynamic origin--destination matrix
  prediction with line graph neural networks and kalman filter,''
  \emph{Transportation Research Record}, vol. 2674, no.~8, pp. 491--503, 2020.

\bibitem{koca2021origin}
D.~Koca, J.~D. Schm{\"o}cker, and K.~Fukuda, ``Origin-destination matrix
  estimation by deep learning using maps with new york case study,'' in
  \emph{2021 7th International Conference on Models and Technologies for
  Intelligent Transportation Systems (MT-ITS)}.\hskip 1em plus 0.5em minus
  0.4em\relax IEEE, 2021, pp. 1--6.

\bibitem{Ye2020}
J.~Ye, J.~Zhao, K.~Ye, and C.~Xu, ``{How to Build a Graph-Based Deep Learning
  Architecture in Traffic Domain: A Survey},'' \emph{IEEE Transactions on
  Intelligent Transportation Systems}, 2020.

\bibitem{liu2021deeptsp}
Y.~Liu, C.~Lyu, Y.~Zhang, Z.~Liu, W.~Yu, and X.~Qu, ``Deeptsp: Deep traffic
  state prediction model based on large-scale empirical data,''
  \emph{Communications in transportation research}, vol.~1, p. 100012, 2021.

\bibitem{Geng2019}
X.~Geng, Y.~Li, L.~Wang, L.~Zhang, Q.~Yang, J.~Ye, and Y.~Liu,
  ``{Spatiotemporal Multi-Graph Convolution Network for Ride-Hailing Demand
  Forecasting},'' \emph{Proceedings of the AAAI Conference on Artificial
  Intelligence}, vol.~33, pp. 3656--3663, 2019.

\bibitem{shi2020predicting}
H.~Shi, Q.~Yao, Q.~Guo, Y.~Li, L.~Zhang, J.~Ye, Y.~Li, and Y.~Liu, ``Predicting
  origin-destination flow via multi-perspective graph convolutional network,''
  in \emph{2020 IEEE 36th International Conference on Data Engineering
  (ICDE)}.\hskip 1em plus 0.5em minus 0.4em\relax IEEE, 2020, pp. 1818--1821.

\bibitem{liu2019attention}
Y.~Liu, Z.~Liu, C.~Lyu, and J.~Ye, ``Attention-based deep ensemble net for
  large-scale online taxi-hailing demand prediction,'' \emph{IEEE transactions
  on intelligent transportation systems}, vol.~21, no.~11, pp. 4798--4807,
  2019.

\bibitem{cao2021bert}
D.~Cao, K.~Zeng, J.~Wang, P.~K. Sharma, X.~Ma, Y.~Liu, and S.~Zhou,
  ``Bert-based deep spatial-temporal network for taxi demand prediction,''
  \emph{IEEE Transactions on Intelligent Transportation Systems}, 2021.

\bibitem{wang2021uncertainty}
W.~Wang and Y.~Wu, ``Is uncertainty always bad for the performance of
  transportation systems?'' \emph{Communications in transportation research},
  vol.~1, p. 100021, 2021.

\bibitem{Zhao2002}
\BIBentryALTinterwordspacing
Y.~Zhao and K.~M. Kockelman, ``{The propagation of uncertainty through travel
  demand models: An exploratory analysis},'' \emph{Annals of Regional Science},
  vol.~36, no.~1, pp. 145--163, 2002. [Online]. Available:
  \url{https://link.springer.com/article/10.1007/s001680200072}
\BIBentrySTDinterwordspacing

\bibitem{qian2023uncertainty}
W.~Qian, D.~Zhang, Y.~Zhao, K.~Zheng, and J.~James, ``Uncertainty
  quantification for traffic forecasting: A unified approach,'' in \emph{2023
  IEEE 39th International Conference on Data Engineering (ICDE)}.\hskip 1em
  plus 0.5em minus 0.4em\relax IEEE, 2023, pp. 992--1004.

\bibitem{wangspatial}
J.~Wang, S.~Guo, T.~Wei, Y.~Zhao, Y.~Lin, H.~Wan \emph{et~al.},
  ``Spatial-temporal uncertainty-aware graph networks for promoting accuracy
  and reliability of traffic forecasting.''

\bibitem{li2020}
C.~Li, L.~Bai, W.~Liu, L.~Yao, and S.~T. Waller, ``{Graph Neural Network for
  Robust Public Transit Demand Prediction},'' \emph{IEEE Transactions on
  Intelligent Transportation Systems}, pp. 1--13, 2020.

\bibitem{maryam2023uncertainty}
H.~Maryam, T.~Panayiotou, and G.~Ellinas, ``Uncertainty quantification and
  consideration in ml-aided traffic-driven service provisioning,''
  \emph{Computer Communications}, vol. 202, pp. 13--22, 2023.

\bibitem{Rodrigues2020}
F.~Rodrigues and F.~C. Pereira, ``{Beyond Expectation: Deep Joint Mean and
  Quantile Regression for Spatiotemporal Problems},'' \emph{IEEE Transactions
  on Neural Networks and Learning Systems}, pp. 1--13, 2020.

\bibitem{zhuang2022uncertainty}
D.~Zhuang, S.~Wang, H.~Koutsopoulos, and J.~Zhao, ``Uncertainty quantification
  of sparse travel demand prediction with spatial-temporal graph neural
  networks,'' in \emph{Proceedings of the 28th ACM SIGKDD Conference on
  Knowledge Discovery and Data Mining}, 2022, pp. 4639--4647.

\bibitem{Pearce2018}
T.~Pearce, M.~Zaki, A.~Brintrup, and A.~Neely, ``{High-quality prediction
  intervals for deep learning: A distribution-free, ensembled approach},''
  \emph{35th International Conference on Machine Learning, ICML 2018}, vol.~9,
  pp. 6473--6482, 2018.

\bibitem{Khosravi2011}
A.~Khosravi, S.~Nahavandi, D.~Creighton, and A.~F. Atiya, ``{Lower upper bound
  estimation method for construction of neural network-based prediction
  intervals},'' \emph{IEEE Transactions on Neural Networks}, vol.~22, no.~3,
  pp. 337--346, 2011.

\bibitem{Nix1994}
D.~A. Nix and A.~S. Weigend, ``{Estimating the mean and variance of the target
  probability distribution},'' in \emph{IEEE International Conference on Neural
  Networks - Conference Proceedings}, 1994.

\bibitem{khosravi2014optimized}
A.~Khosravi and S.~Nahavandi, ``An optimized mean variance estimation method
  for uncertainty quantification of wind power forecasts,'' \emph{International
  Journal of Electrical Power \& Energy Systems}, vol.~61, pp. 446--454, 2014.

\bibitem{Koenker2001}
R.~Koenker and K.~F. Hallock, ``{Quantile regression},'' \emph{Journal of
  Economic Perspectives}, 2001.

\bibitem{Kabir2018}
H.~M. Kabir, A.~Khosravi, M.~A. Hosen, and S.~Nahavandi, ``{Neural
  Network-Based Uncertainty Quantification: A Survey of Methodologies and
  Applications},'' \emph{IEEE Access}, vol.~6, pp. 36\,218--36\,234, 2018.

\bibitem{Gawlikowski2021}
\BIBentryALTinterwordspacing
J.~Gawlikowski, C.~R.~N. Tassi, M.~Ali, J.~Lee, M.~Humt, J.~Feng, A.~Kruspe,
  R.~Triebel, P.~Jung, R.~Roscher, M.~Shahzad, W.~Yang, R.~Bamler, and X.~X.
  Zhu, ``{A Survey of Uncertainty in Deep Neural Networks},'' jul 2021.
  [Online]. Available: \url{https://arxiv.org/abs/2107.03342v3}
\BIBentrySTDinterwordspacing

\bibitem{Blundell2015}
C.~Blundell, J.~Cornebise, K.~Kavukcuoglu, and D.~Wierstra, ``{Weight
  uncertainty in neural networks},'' \emph{32nd International Conference on
  Machine Learning, ICML 2015}, vol.~2, pp. 1613--1622, 2015.

\bibitem{Gal2016}
Y.~Gal and Z.~Ghahramani, ``{Dropout as a Bayesian approximation: Representing
  model uncertainty in deep learning},'' in \emph{33rd International Conference
  on Machine Learning, ICML 2016}, 2016.

\bibitem{Sicking2021}
\BIBentryALTinterwordspacing
J.~Sicking, M.~Akila, M.~Pintz, T.~Wirtz, A.~Fischer, and S.~Wrobel, ``{A Novel
  Regression Loss for Non-Parametric Uncertainty Optimization},'' pp.
  2021--2022, jan 2021. [Online]. Available:
  \url{https://arxiv.org/abs/2101.02726v1 http://arxiv.org/abs/2101.02726}
\BIBentrySTDinterwordspacing

\bibitem{Heskes1997}
T.~Heskes, ``{Practical confidence and prediction intervals},'' in
  \emph{Advances in Neural Information Processing Systems}, 1997.

\bibitem{Lakshminarayanan2016}
\BIBentryALTinterwordspacing
B.~Lakshminarayanan, A.~Pritzel, and C.~Blundell, ``{Simple and Scalable
  Predictive Uncertainty Estimation using Deep Ensembles},'' \emph{31st
  Conference on Neural Information Processing Systems (NIPS 2017)}, vol.~43,
  no.~2, pp. 145--150, dec 2016. [Online]. Available:
  \url{http://arxiv.org/abs/1612.01474}
\BIBentrySTDinterwordspacing

\bibitem{Chen2020}
\BIBentryALTinterwordspacing
E.~Chen, Z.~Ye, C.~Wang, and M.~Xu, ``{Subway Passenger Flow Prediction for
  Special Events Using Smart Card Data},'' \emph{IEEE Transactions on
  Intelligent Transportation Systems}, vol.~21, no.~3, pp. 1109--1120, mar
  2020. [Online]. Available:
  \url{https://ieeexplore.ieee.org/document/8725576/}
\BIBentrySTDinterwordspacing

\bibitem{Guo2014}
\BIBentryALTinterwordspacing
J.~Guo, W.~Huang, and B.~M. Williams, ``{Adaptive Kalman filter approach for
  stochastic short-term traffic flow rate prediction and uncertainty
  quantification},'' \emph{Transportation Research Part C: Emerging
  Technologies}, vol.~43, pp. 50--64, 2014. [Online]. Available:
  \url{http://dx.doi.org/10.1016/j.trc.2014.02.006}
\BIBentrySTDinterwordspacing

\bibitem{Petrik2016}
\BIBentryALTinterwordspacing
O.~Petrik, F.~Moura, and J.~d. A.~e. Silva, ``{Measuring uncertainty in
  discrete choice travel demand forecasting models},'' \emph{Transportation
  Planning and Technology}, vol.~39, no.~2, pp. 218--237, feb 2016. [Online].
  Available:
  \url{https://www.tandfonline.com/action/journalInformation?journalCode=gtpt20}
\BIBentrySTDinterwordspacing

\bibitem{Cools2010}
\BIBentryALTinterwordspacing
M.~Cools, B.~Kochan, T.~Bellemans, D.~Janssens, and G.~Wets, ``{Assessment of
  the effect of micro-simulation error on key travel indices: evidence from the
  activity-based model feathers},'' \emph{90th Annual Meeting of the
  Transportation Research Board}, pp. 1--15, 2010. [Online]. Available:
  \url{https://orbi.uliege.be/handle/2268/134326}
\BIBentrySTDinterwordspacing

\bibitem{vlahogianni2011temporal}
E.~Vlahogianni and M.~Karlaftis, ``Temporal aggregation in traffic data:
  implications for statistical characteristics and model choice,''
  \emph{Transportation Letters}, vol.~3, no.~1, pp. 37--49, 2011.

\bibitem{zhang2013univariate}
Y.~Zhang, R.~Sun, A.~Haghani, and X.~Zeng, ``Univariate volatility-based models
  for improving quality of travel time reliability forecasting,''
  \emph{Transportation research record}, vol. 2365, no.~1, pp. 73--81, 2013.

\bibitem{Kipf2017}
T.~N. Kipf and M.~Welling, ``{Semi-supervised classification with graph
  convolutional networks},'' \emph{5th International Conference on Learning
  Representations, ICLR 2017 - Conference Track Proceedings}, pp. 1--14, 2017.

\bibitem{velivckovic2017graph}
P.~Veli{\v{c}}kovi{\'c}, G.~Cucurull, A.~Casanova, A.~Romero, P.~Lio, and
  Y.~Bengio, ``Graph attention networks,'' \emph{arXiv preprint
  arXiv:1710.10903}, 2017.

\bibitem{Kiureghian2009}
\BIBentryALTinterwordspacing
A.~D. Kiureghian and O.~Ditlevsen, ``{Aleatory or epistemic? Does it matter?}''
  \emph{Structural Safety}, vol.~31, no.~2, pp. 105--112, 2009. [Online].
  Available: \url{http://dx.doi.org/10.1016/j.strusafe.2008.06.020}
\BIBentrySTDinterwordspacing

\bibitem{ye2021sparse}
Y.~Ye and S.~Ji, ``Sparse graph attention networks,'' \emph{IEEE Transactions
  on Knowledge and Data Engineering}, 2021.

\end{thebibliography}
\begin{IEEEbiography}[{\includegraphics[width=1in,height=1.25in,clip,keepaspectratio]{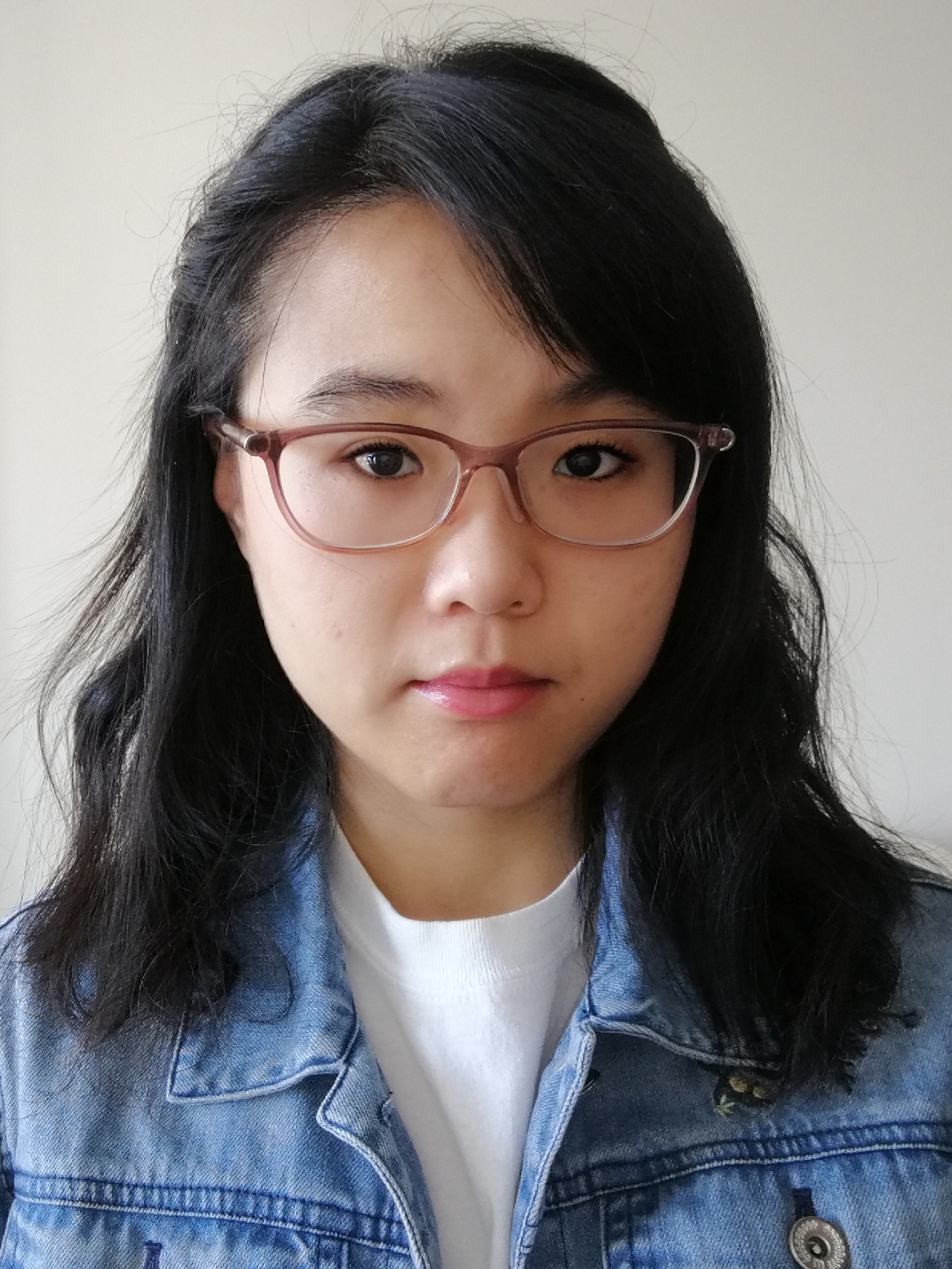}}]{Qingyi Wang} is a PhD student in transportation. She graduated with a Bachelor’s of Applied Science in Engineering Science from the University of Toronto in 2018, and with a Master of Science in Transportation from MIT in 2020. Her research focuses on bringing machine learning into transportation demand modelling, and making use of domain knowledge to improve and better interpret machine learning models. 

\end{IEEEbiography}

\begin{IEEEbiography}[{\includegraphics[width=1in,height=1.25in,clip,keepaspectratio]{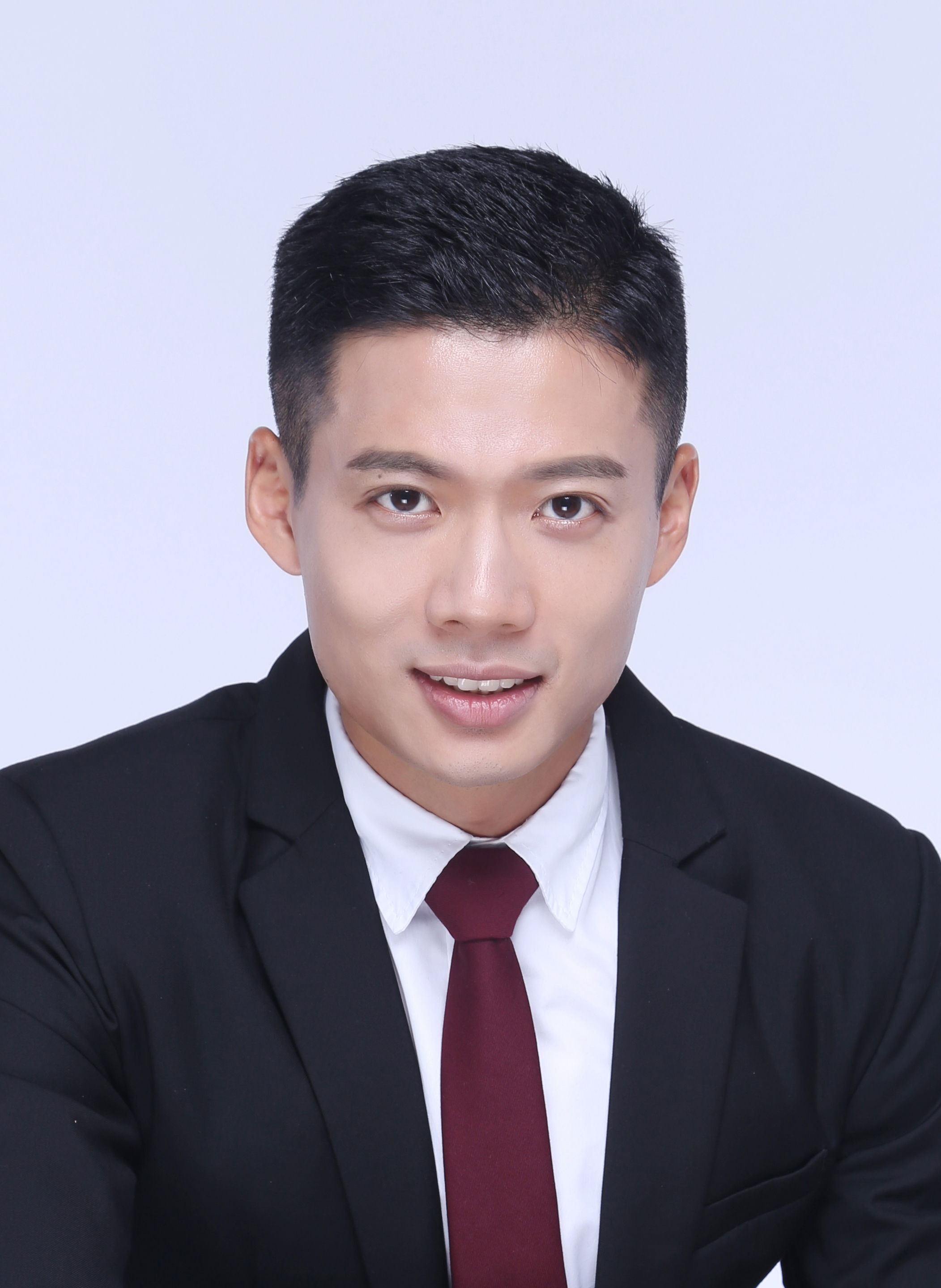}}]{Shenhao Wang}
is an assistant professor at the University of Florida and research affiliate at MIT Urban Mobility Lab and Human Dynamics Group in Media Lab. His research focuses on developing interpretable, generalizable, and ethical deep learning models to analyze individual decision-making with applications to urban mobility. He synergizes discrete choice models and deep neural networks in travel demand modeling by interpreting the “black box” deep neural networks with economic theory. Shenhao Wang completed his interdisciplinary Ph.D. in Computer and Urban Science from MIT in 2020.
\end{IEEEbiography}

\begin{IEEEbiography}[{\includegraphics[width=1in,height=1.25in,clip,keepaspectratio]{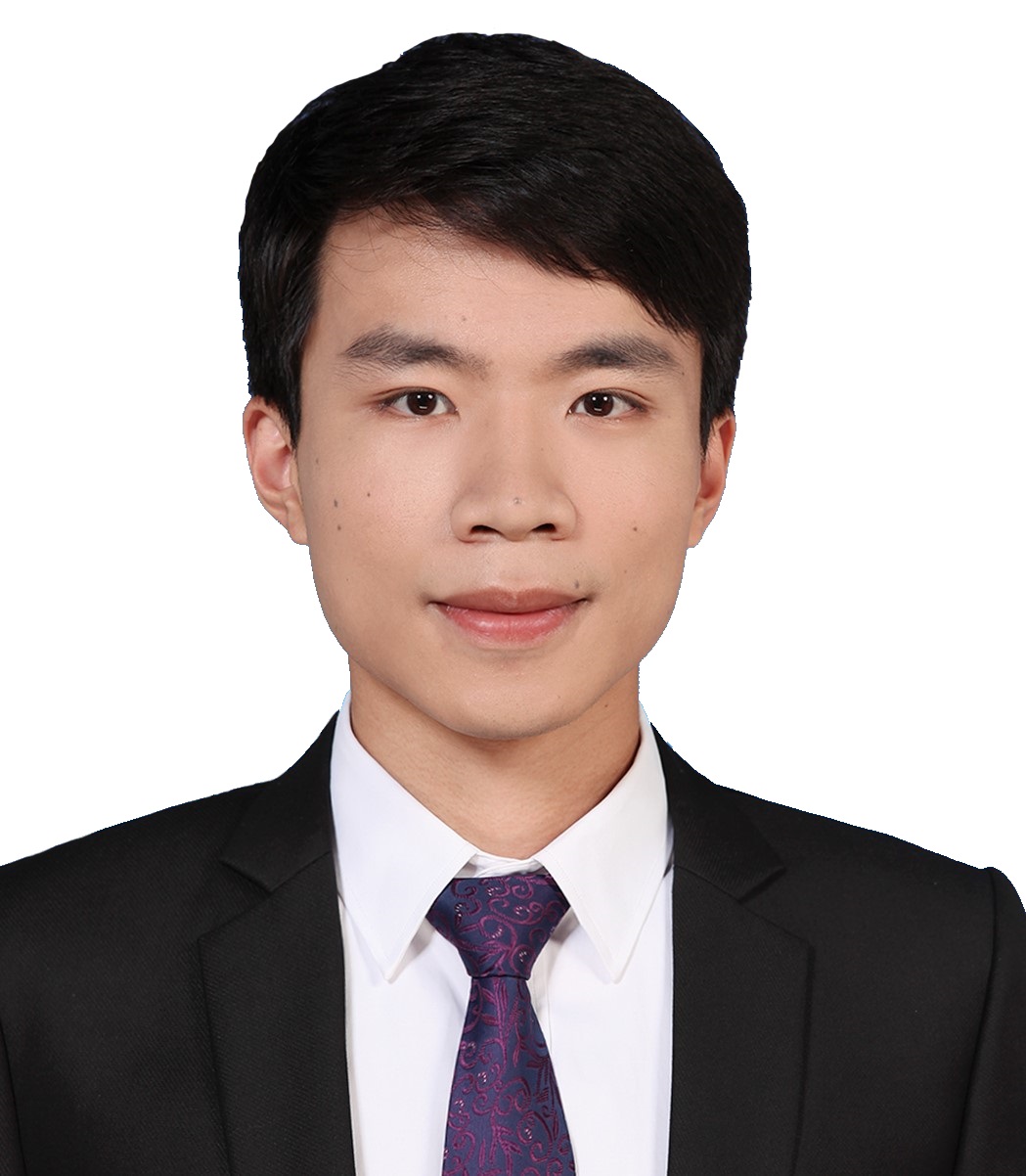}}]{Dingyi Zhuang} is a master student at MIT Urban Mobility Lab. He received a B.S. degree in Mechanical Engineering from Shanghai Jiao Tong University in 2019, and an M.Eng. degree in Transportation Engineering from McGill University, in 2021. His research interests lie in deep learning, urban computing, and spatiotemporal data modeling.
\end{IEEEbiography}

\begin{IEEEbiography}[{\includegraphics[width=1in,height=1.25in,clip,keepaspectratio]{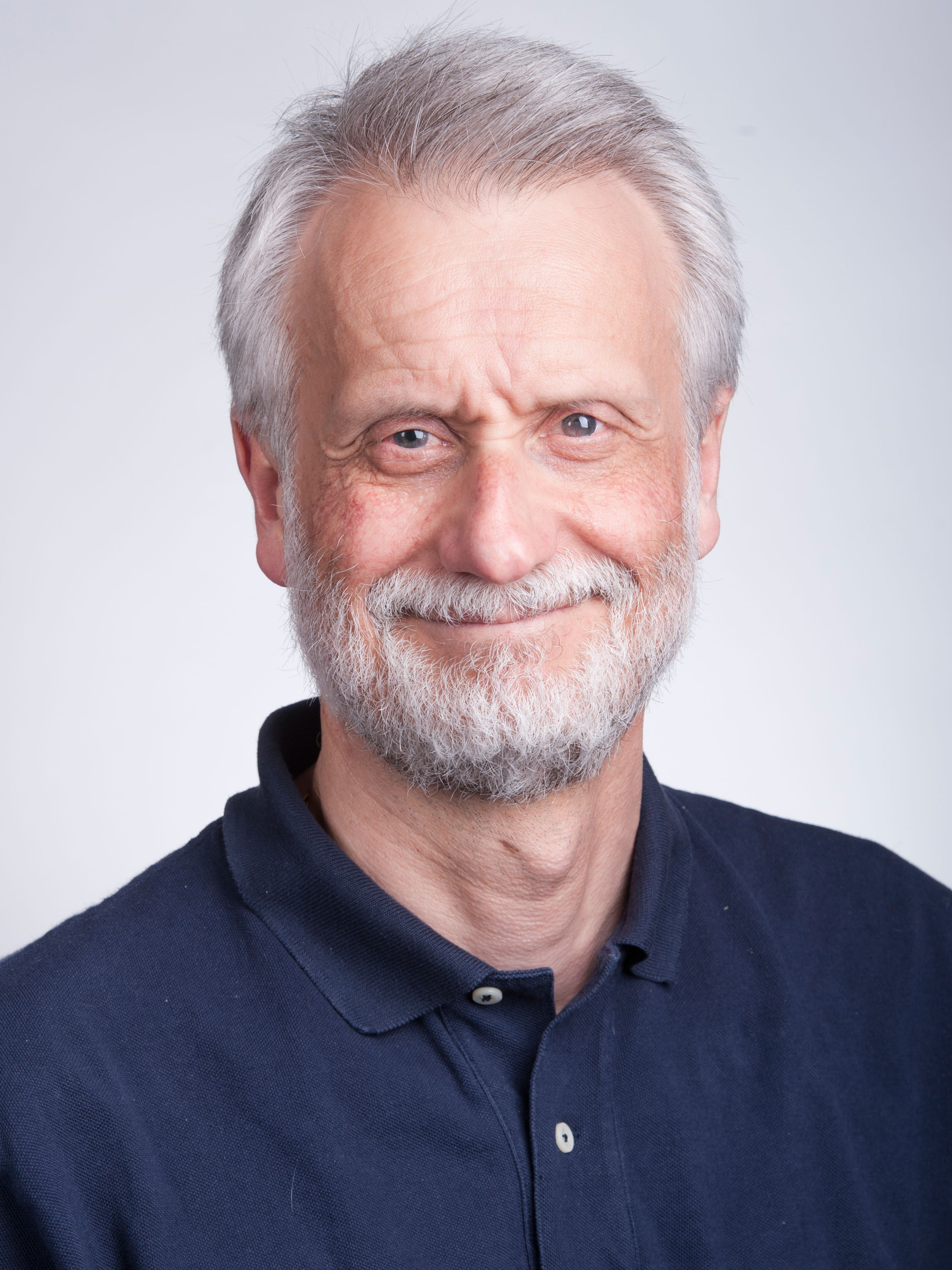}}]{Haris Koutsopoulos}
is currently a Professor with the Department of Civil and Environmental Engineering, Northeastern University, Boston, MA, USA, and a Guest Professor with the KTH Royal Institute of Technology, Stockholm. He founded the iMobility Laboratory, which uses information and communication technologies to address urban mobility problems. His current research focuses on the use of data from opportunistic and dedicated sensors to improve planning, operations, monitoring, and control of urban transportation systems, including public transportation. The laboratory received the IBM Smarter Planet Award in 2012.
\end{IEEEbiography}


\begin{IEEEbiography}[{\includegraphics[width=1in,height=1.25in,clip,keepaspectratio]{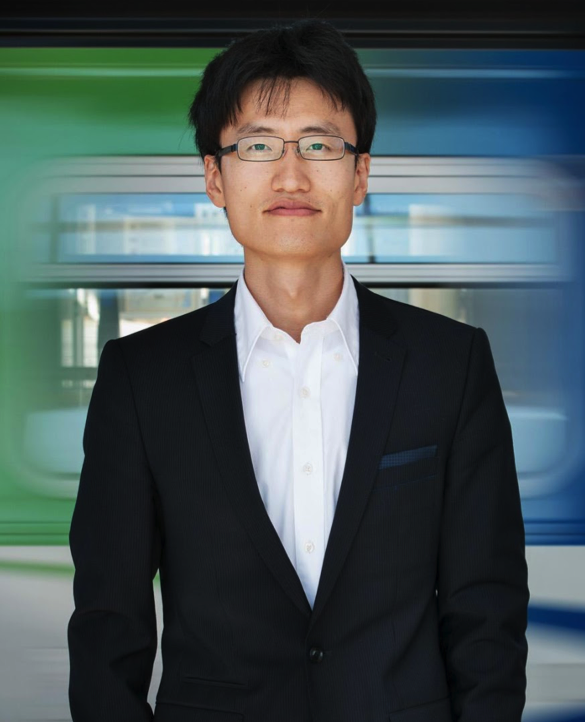}}]{Jinhua Zhao}
is currently the Edward H. and Joyce Linde Associate Professor of city and transportation planning at MIT. He brings behavioral science and transportation technology together to shape travel behavior, design mobility systems, and reform urban policies. He directs the MIT Urban Mobility Laboratory and Public Transit Laboratory.
\end{IEEEbiography}

\end{document}